\documentclass[lettersize,journal]{IEEEtran}
\usepackage{amsmath,amsfonts}
\usepackage{algorithmic}
\usepackage[ruled]{algorithm2e}
\usepackage{xcolor,soul,framed} 

\colorlet{shadecolor}{yellow}
\usepackage{color,soul}
\usepackage[pdftex]{graphicx}

\usepackage{array}
\usepackage[hyphens]{url}
\usepackage{multirow}
\usepackage{amsmath}
\usepackage{threeparttable} 

\usepackage{subfigure}
\usepackage{graphicx}
\usepackage{multirow}
\usepackage{hyperref}
\usepackage{bbm}
\usepackage{bm}

\newtheorem{remark}{Remark}

\begin{document}

\title{Bilevel Multi-Armed Bandit-Based Hierarchical Reinforcement Learning for Interaction-Aware Self-Driving at Unsignalized Intersections}

\author{
Zengqi Peng, Yubin Wang, Lei Zheng,
and Jun Ma, \textit{IEEE, Senior Member}
  \thanks{This work was supported by the National Natural Science Foundation of China under Grant 62303390. \textit{(Corresponding author: Jun Ma.)}}
\thanks{Zengqi Peng, Yubin Wang, and Lei Zheng are with the Robotics and Autonomous Systems Thrust, The Hong Kong University of Science and Technology (Guangzhou), Guangzhou 511453, China
(e-mail: zpeng940@connect.hkust-gz.edu.cn; ywang575@connect.hkust-gz.edu.cn; lzheng135@connect.hkust-gz.edu.cn).
}
\thanks{Jun Ma is with the Robotics and Autonomous Systems Thrust, The Hong Kong University of Science and Technology (Guangzhou), Guangzhou 511453, China, and also with the Division of Emerging Interdisciplinary Areas, The Hong Kong University of Science and Technology, Hong Kong SAR, China. (e-mail: jun.ma@ust.hk).} 

}

\maketitle

\begin{abstract}
In this work, we present BiM-ACPPO, a bilevel multi-armed bandit-based hierarchical reinforcement learning framework for interaction-aware decision-making and planning at unsignalized intersections. Essentially, it proactively takes the uncertainties associated with surrounding vehicles (SVs) into consideration, which encompass those stemming from the driver's intention, interactive behaviors, and the varying number of SVs. Intermediate decision variables are introduced to enable the high-level RL policy to provide an interaction-aware reference, for guiding low-level model predictive control (MPC) and further enhancing the generalization ability of the proposed framework. By leveraging the structured nature of self-driving at unsignalized intersections, the training problem of the RL policy is modeled as a bilevel curriculum learning task, which is addressed by the proposed Exp3.S-based BiMAB algorithm. It is noteworthy that the training curricula are dynamically adjusted, thereby facilitating the sample efficiency of the RL training process. Comparative experiments are conducted in the high-fidelity CARLA simulator, and the results indicate that our approach achieves superior performance compared to all baseline methods. Furthermore, experimental results in two new urban driving scenarios clearly demonstrate the commendable generalization performance of the proposed method. 

\end{abstract}


\section{Introduction}

Recently, the autonomous driving community has observed significant progress and impressive achievements across both academic research and industry applications \cite{xing2019driver,mozaffari2020deep,lu2023game}. 
In contrast to highway driving tasks, the challenges in urban driving tasks primarily arise from the complexity of the environment, including uncertain intentions among surrounding vehicles (SVs), frequent interactions between vehicles, and the diversity of road layouts \cite{paden2016survey,zheng2024spatiotemporal}. 
Apparently, inaccurate assessments regarding driving intentions of SVs, such as yielding or not, could significantly potentially lead to traffic accidents and congestion, thereby hindering the safe and efficient motion of the ego vehicle (EV) \cite{nan2022intention}. 
This is particularly evident at intersections, where vehicles originating from diverse directions are required to swiftly traverse a shared central zone to access their respective target lanes without collisions. 
The situation becomes even more severe in unsignalized intersections which generally exhibit unpredictable traffic patterns. Because the EV is required to interact with a varying number of SVs arriving from diverse directions simultaneously in the absence of traffic signals. 
Uncertainties arising from the driving maneuvers and intentions of SVs present serious safety and efficiency concerns in the interaction process. This requires enhancing the capabilities of the EV to prevent potential risks. 
In this sense, self-driving at unsignalized intersections requires advanced decision-making ability and interaction awareness to interact safely and effectively with SVs exhibiting multi-modal behaviors.

Rule-based and optimization-based methods are two representative approaches for autonomous driving at intersections, both of which have been extensively studied. 
In general, rule-based methods predefine a series of rules about potential traffic situations for autonomous vehicles to select suitable behaviors from the rule base. 
A set of rules is designed to regulate the order of vehicles to pass through the unsignalized intersection complying with the law of road traffic safety \cite{lu2014rule,aksjonov2021rule}. 
Nevertheless, it is difficult for rule-based approaches to encompass the entire spectrum of traffic situations, typically tending to conservative driving behaviors \cite{tian2020game}. This could cause traffic congestion and compromise traffic efficiency. 
On the other hand, optimization-based methods formulate the self-driving tasks as mathematical optimization problems that minimize a well-designed cost function subject to a set of constraints \cite{riegger2016centralized,kneissl2018feasible,huang2023decentralized}. 
A two-stage optimization approach is proposed for self-driving tasks at unsignalized intersections by constructing the crossing problem as a mixed-integer linear programming and linear programming \cite{yao2023two}. Considering the existence of non-cooperative vehicles, a differential game-based optimization strategy is proposed to make the controlled vehicle coordinate with non-cooperative SVs \cite{huang2024non}. 
However, these methods are computationally expensive and time-consuming, and it is difficult to consider unexpected or rapidly changing traffic scenarios. 
Furthermore, optimization-based approaches tend to struggle with poor task efficiency in complex environments, particularly those involving SVs with diverse interactive behaviors. This could result in decisions that are potentially hazardous for autonomous vehicles.

Reinforcement learning (RL) technologies have recently shown significant potential in propelling the development of the autonomous driving community \cite{shu2021driving,qiao2021behavior,liao2023integration}. A multi-task DQN algorithm is proposed to control the speed of vehicles to navigate through intersections \cite{kai2020multi}, where the intersection navigation task is decomposed into several sub-tasks for the construction of the reward function. However, it does not consider the difficulty and potential collision risks of different types of sub-tasks. 
A social attention module is incorporated into the RL framework to balance the safety and efficiency of self-driving at intersections \cite{liu2023multi}. 
While RL-based methods have achieved impressive results at intersection scenarios, 
most of these works do not consider vehicle dynamics and driving comfort. 
To tackle this issue, a potential solution is to integrate the RL with optimization-based methods represented by model predictive control (MPC) \cite{song2022policy,wang2023chance}. In this sense, the MPC can consider state and control constraints under the decisions provided by RL. 
This solution combines their advantages to establish a hierarchical framework. 
A hybrid decision-making algorithm is designed for autonomous vehicles at intersections \cite{tram2019learning}, where high-level decisions are derived for the reference of a low-level planning module. However, its action space only takes several basic behaviors, which could lead to poor maneuverability. 
To enhance the maneuverability of the EV, an intuitive action space, consisting of velocity and heading angle, is utilized in \cite{al2023self} for unprotected left-turn tasks. 
However, the driving intention of the SVs is assumed to not decelerate or yield to the EV, 
which could limit the adaptability of the trained policy in diverse situations, such as SV driving towards different lanes and giving way to the EV. 
Furthermore, this work solely considers left-turn tasks at unsignalized intersections, which could limit the generalization ability of the trained policy in other types of tasks within unsignalized intersections, such as go-straight and right-turn tasks. Besides, go-straight tasks at intersections present multiple potential collision points, thereby posing a challenge to self-driving vehicles.

Additionally, one significant limitation of the RL module within the hierarchical methods is the poor sample efficiency when it comes to complex driving environments \cite{huang2023efficient}. 
The RL agent necessitates considerable training episodes to achieve satisfactory performance due to the randomness of exploration. 
Therefore, 
it would be challenging for RL agents to thoroughly explore the environment and learn effective policies within limited resources if without suitable guidance. 
To address this problem, the environment model is incorporated into the RL algorithm to enhance the sample efficiency, where virtual samples are generated to accelerate the training process \cite{guan2020centralized}. However, the effectiveness of the trained policy is heavily dependent on the accuracy of the model. The performance of the policy will be severely compromised if there is a significant discrepancy between the environmental model and the actual environment. 
Additionally, directly training the RL agent within complex environments could result in a solution with limited driving performance. 
Curriculum learning technique provides a promising way to alleviate the above problems \cite{bengio2009curriculum,song2021autonomous}. 
In \cite{peng2023CPPO}, a stage-decaying curriculum learning approach is proposed to guide the training of the RL policy. 
However, since the scheduling of curriculum shifts is predetermined manually, the effectiveness of the training results relies on expert experience heavily.

To address the above issues, diverse automated curriculum learning methods have been proposed \cite{qiao2018automatically,khaitan2022state}. 
An adaptive curriculum-based method is utilized for training the RL policies at single-lane intersections, which models various initial locations of the EV as curriculum features \cite{qiao2018automatically}. 
However, the adjustment in the importance of curriculum depends on their respective weights, which could result in delayed or even potentially failed transitions to curricula with promising high rewards in the future. 
In \cite{khaitan2022state}, a state drop-out curriculum learning method is developed to learn an RL policy at unsignalized intersections. 
It incrementally trains the RL policy by omitting information about future states throughout the training process. 
Nevertheless, these works assume that SVs will not respond to the behavior of the EV, and their future trajectories are known to the EV. These simplifications could jeopardize driving safety and degrade the generalization ability of RL policies. 
It is worth noting that human drivers would demonstrate diverse driving maneuvers according to the behaviors of SVs in the real world, which is crucial for achieving safe and efficient interactions. 
An interactive planning framework is proposed for the behavior generation of EV in an intersection scenario \cite{xia2022interactive}. However, this work only considers the intentions related to the task objectives of SVs while ignoring their responses to the behaviors of the EV. Besides, the simple action space impedes the flexible behaviors of the EV and also the generalization ability into various driving scenarios. 
Essentially, as most current studies do not comprehensively consider the interactive behaviors of SVs, these simplifications could limit the applicability of autonomous driving techniques in real-world scenarios.

Therefore, this paper presents a novel bilevel multi-armed bandit-based automated curriculum PPO (BiM-ACPPO) framework, which integrates a bilevel multi-armed bandit (BiMAB) module and a hierarchical RL framework, to improve training efficiency and driving performance. 
Through the dynamic assessment and adjustment of the training process, the RL agent will progressively enhance the self-driving strategy at unsignalized intersections, starting from simple to more challenging scenarios. 
The main contributions are listed as follows:
\vspace{-0.1cm}
\begin{itemize}
    \item A novel BiM-ACPPO framework is proposed for interaction-aware decision-making and planning at unsignalized intersections, which can proactively handle the uncertainty stemming from the driving intentions of SVs with multi-modal interaction behaviors and traffic density. A special action space is utilized to enhance the generalization ability of the framework.

    \item  A Exp3.S-based BiMAB algorithm is devised for automated curriculum selection by leveraging the structured nature of the self-driving tasks at unsignalized intersections, thereby facilitating efficient sampling and effective exploration in the RL training process. 
    
    \item  We demonstrate the effectiveness of the proposed approach in the high-fidelity simulator CARLA. 
    The BiMAB exhibits appropriate curriculum transition timing during the training process, and the proposed framework achieves superior performance and commendable generalization ability compared to all baseline methods. 

\end{itemize}

The rest of the paper is structured as follows. Section II presents the problem statement. Section III illustrates the proposed BiMAB approach and hierarchical RL framework. Section IV demonstrates the experimental results followed by pertinent analysis. Finally, Section V summarizes the conclusion and discusses future works.

\section{Problem Statement}
\label{Section2}
\subsection{Problem Statement}

\begin{figure}[!htbp] 
    \centering   
    \includegraphics[trim=0 0cm 0 0.5cm, width=0.8\linewidth]{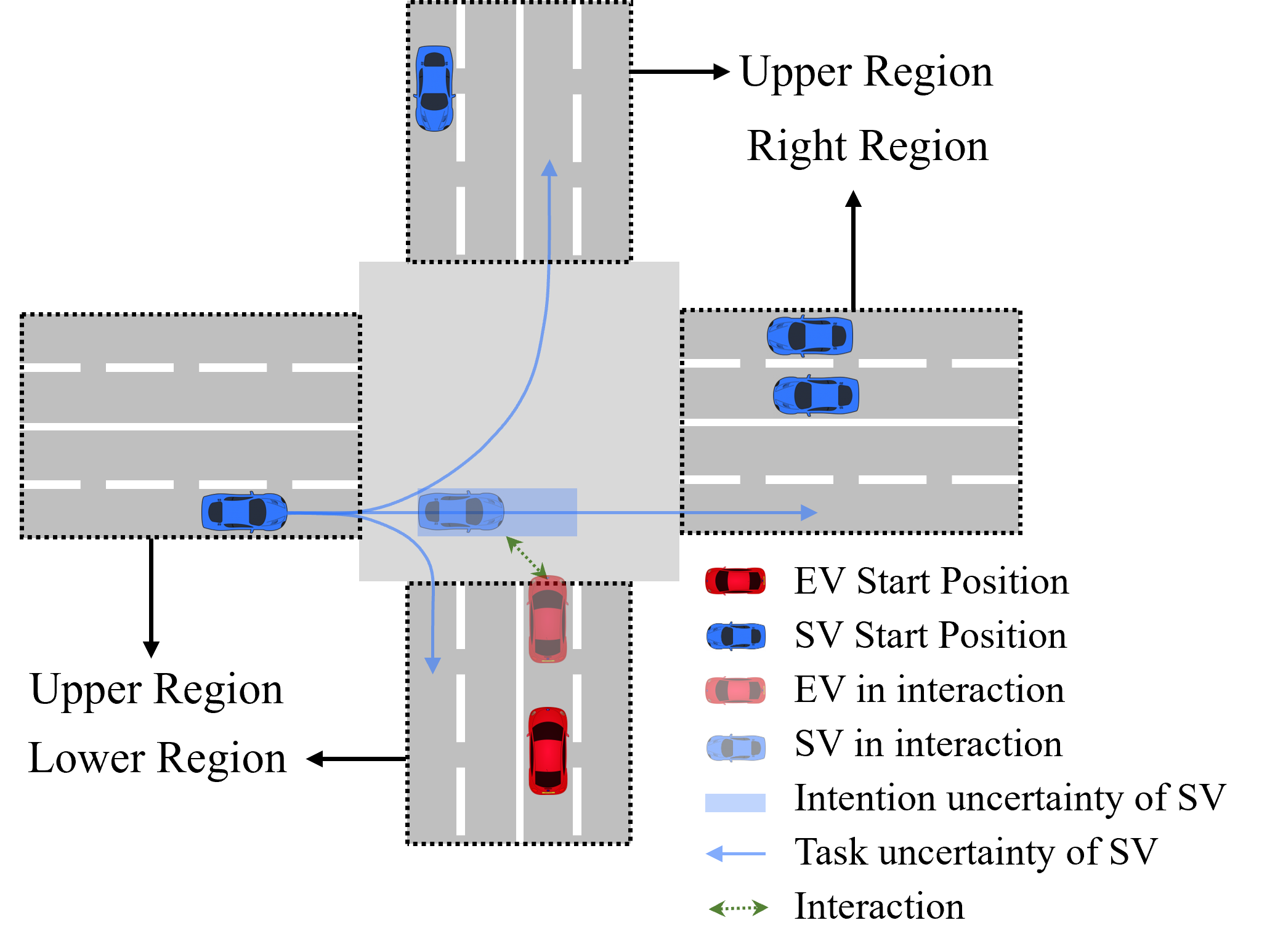}
   \caption{Overview of the task scenario. The EV is illustrated in red, while SVs are illustrated in blue. The uncertainties include the task and intention uncertainty of SVs and the varying number of SVs. All SVs will respond to the behavior of other vehicles.
   }
    \label{target_scene}
\end{figure}

In this work, the task scenario is set as a dual-lane unsignalized intersection, which is shown in Fig. \ref{target_scene}. The start position and target position of the EV are generated randomly within the lower region and the other three regions, respectively, obeying the traffic rules. A random number of SVs with diverse interactive styles start from random positions of the upper, left, and right regions, and they drive towards respective target lanes simultaneously. The goal is to generate a control action sequence to guide the EV to safely and swiftly complete diverse crossing tasks, including unprotected left-turn, go-straight, and right-turn tasks.

In contrast with \cite{qiao2018automatically} and \cite{khaitan2022state}, the behaviors of all vehicles within the environment are interdependent, and different SVs exhibit diverse driving styles, including aggressive, moderate, and conservative styles. For example, SVs could either yield or disregard potential collision risks and continue driving when interacting with other vehicles at conflict points or potential collision points. Hence, the SVs in the target scenarios demonstrate multi-modal and interactive behaviors. 
Additionally, the number of SVs is also not fixed in diverse scenario configurations. In this context, the EV is required to complete various crossing tasks in the environment with the interactive SVs. 
Here, we assume that the EV can access the position and velocity information of SVs, while their goal tasks and driving intentions are unknown. 
These configurations inject a significant level of randomness into the driving scenarios, rendering the tasks challenging but close to real-world situations.

\subsection{Vehicle Model}
 
Since the EV drives at low speed in the task scenarios, the dynamic impact of the vehicle is negligible. Therefore, this work utilizes the bicycle model \cite{eiras2021two} to describe the dynamics of the EV, which is expressed as follows:
    \begin{equation}
    \dot{\mathbf{x}}=f(\mathbf{x}, \mathbf{u})=
    \left[\begin{array}{c}
    v \cos (\psi+\delta) \\
    v \sin (\psi+\delta) \\
    \frac{2v}{L} \sin \delta \\
    a 
    \end{array}\right],
    \label{eq:bicycle}
    \end{equation}
where $\mathbf{x}= \begin{bmatrix}  x & y & v & \psi \end{bmatrix} ^T$ is the state vector of the vehicle 
in the global coordinate $\mathcal{W}_g$; $x$ and $y$ denote the X-coordinate and Y-coordinate position of the center of the vehicle, respectively; $\psi$ and $v$ represent the heading angle and the speed, respectively; $\mathbf{u}= \begin{bmatrix} a & \delta \end{bmatrix} ^T$ is the control input vector; $a$ and $\delta$ are the acceleration and steering angle, respectively; $L$ is the inter-axle distance of the vehicle.

\subsection{Model Predictive Control}

First, we discretize (\ref{eq:bicycle}) into the following model:
\begin{equation}
    \mathbf{x}_{k+1} = \mathbf{x}_{k} + f(\mathbf{x}_{k},\mathbf{u}_{k}) \cdot d_k,
\end{equation}
where $\mathbf{x}_{k}$ and $\mathbf{u}_{k}$ represent the state vector and the control input vector at the time step $k$; $f$ is the dynamic function; $d_k$ is the sampling interval.
Let $\mathbf{x}_g$ represent the goal state vector for the EV, which consists of the target location and orientation. The MPC aims to force the EV system to approach the goal state vector by solving a standard optimization problem as follows:
\begin{equation}
\begin{aligned}
&\min _{\mathbf{x}_{1: N}, \mathbf{u}_{0: N-1}} \mathcal{J} = J(\mathbf{x}_N,\mathbf{x}_g) + \sum^N J(\mathbf{x}_k,\mathbf{u}_k,\mathbf{x}_g),\\
&\text {\quad \quad s.t.} \quad \mathbf{x}_{k+1}=\mathbf{x}_k+f\left(\mathbf{x}_k, \mathbf{u}_k\right) d_k,\\
& \quad \quad \quad \quad \mathcal{G}(\mathbf{x},\mathbf{u})\leq 0,\\
& \quad \quad \quad \quad \mathcal{H}(\mathbf{x},\mathbf{u})= 0,\\
& \quad \quad \quad \quad \mathbf{x}_0 = \mathbf{x}({k}),
\end{aligned}
\label{mpc_org}
\end{equation}
where $J(\mathbf{x}_N,\mathbf{x}_g)$ and $J(\mathbf{x}_k,\mathbf{u}_k,\mathbf{x}_g)$ represent terminal cost and running cost, respectively; $N$ is the length of the horizon; $\mathcal{G}(\mathbf{x},\mathbf{u})\leq 0$ and $\mathcal{H}(\mathbf{x},\mathbf{u})=0$ are inequality constraints and equality constraints, respectively; $x_0$ is the initial state of the controlled system in the MPC problem. 
The control objective can be achieved by repeating the solving process and execution in real-time until the system reaches the goal state $\mathbf{x}_g$. 
For complex and dynamic environments, MPC with fixed parameters could encounter issues such as high computational burdens, compromised driving safety, and poor adaptability. In this work, novel high-level decision variables are introduced for parameterization of the MPC formulation. Then the output of the MPC can be modulated to generate diverse behaviors for the EV to interact with multi-model SVs by leveraging the decoded value of the high-level decision variables. 

\begin{figure*}[!t] 
    \centering   
    \includegraphics[trim=0 0cm 0 0.5cm, width=0.88\linewidth]{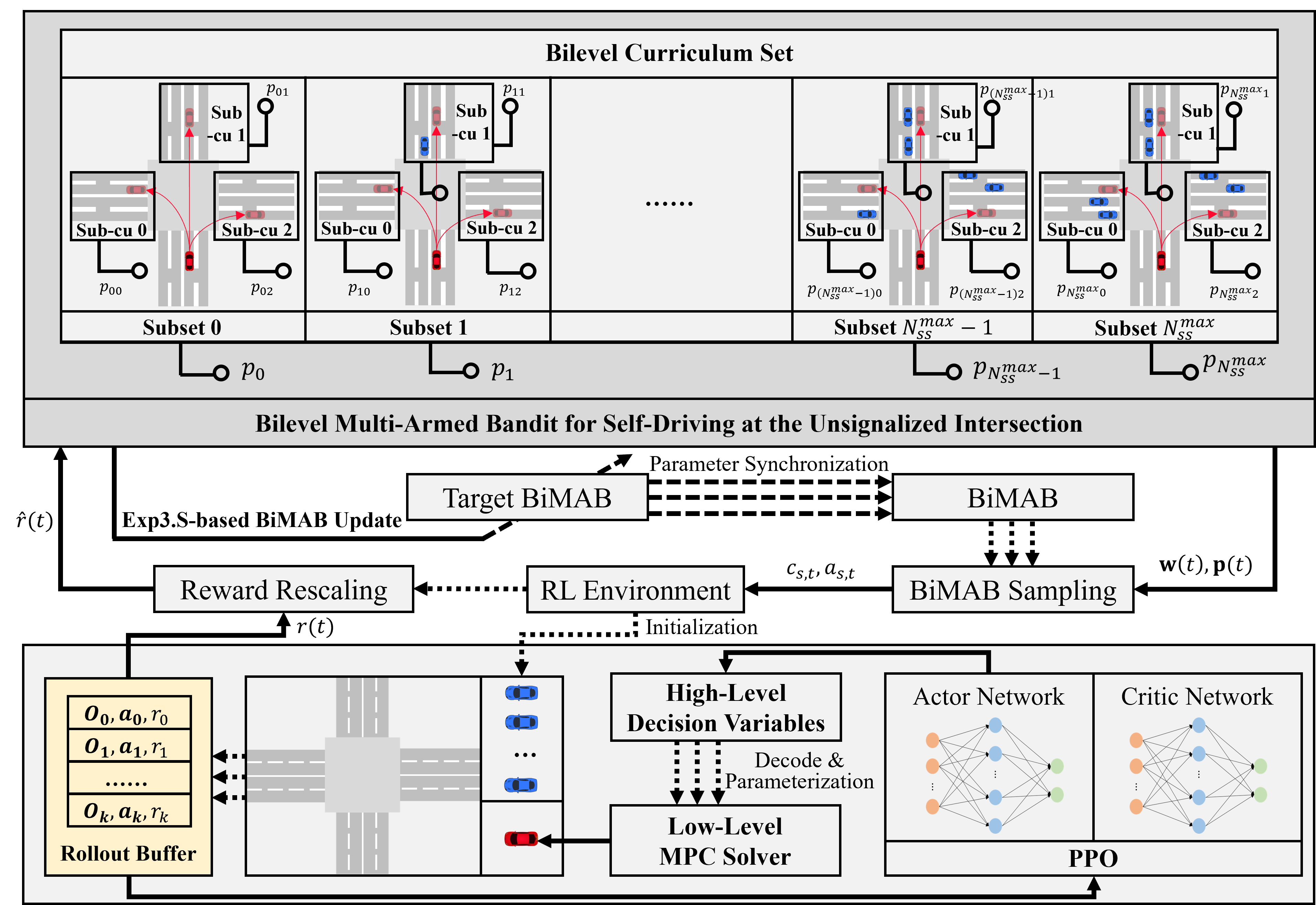}
   \caption{Overview of the BiM-ACPPO approach for interaction-aware self-driving at unsignalized intersections with interactive SVs. The EV and SVs are illustrated in red and blue, respectively. 
  The proposed BiMAB framework models the training process as a bilevel clustered structure. 
  The variables within the first layer and the second layer, are the number of SVs and the task type of the EV, respectively. The solid car and the semi-transparent cars within the BiMAB module denote the start position and the target position of the EV, respectively. 
   }
    \label{frame:whole}
\end{figure*}

\subsection{Learning Environment}

In this work, the target tasks can be modeled as a Markov Decision Process (MDP), where all vehicles are located near the intersection. 
Here, we express the MDP as a tuple $\mathcal{E} = \langle \mathcal{S}, \mathcal{A}, \mathcal{P}, \mathcal{R}, \gamma \rangle$, where elements $\mathcal{S}, \mathcal{A}, \mathcal{P}, \mathcal{R}, \gamma$ are defined as follows:

\textbf{State space $\mathcal{S}$}: In this work, $\mathcal{S}$ comprises kinematic features of EV and SVs within the observation range of the EV. The state matrix at time step $k$ is defined as follows:
\begin{equation}
\begin{split}
\mathbf{S}_{k}=\left[\ \mathbf{s}_{k}^0\ \ \mathbf{s}_{k}^1\ ...\ \mathbf{s}_{k}^{N_{\text{sv}}^{\max}}\ \right]^T,
\end{split}
\label{state martix}
\end{equation}
where $N_{\text{sv}}^{\max}$ denotes the maximum number of SVs; $\mathbf{s}_{k}^0$ and $\mathbf{s}_{k}^i\ (i=1,2,...,N_{\text{sv}}^{\max})$ represent the state of the EV and the state of the $i$-th SV, respectively. 
Specifically, $\mathbf{s}_{k}^i$ is defined as follows:
\begin{equation}
\begin{split}
\mathbf{s}_{k}^{i}={\left[\begin{array}{l l l l l l l}{x_{k}^{i}}&{y_{k}^{i}}&{v_{k}^{i}}&{\psi_{k}^{i}}\end{array}\right]}^{T},
\end{split}
\label{state_space}
\end{equation}
where $x_{k}^{i}$ and $y_{k}^{i}$ represent the X-axis and Y-axis coordinates of the vehicle $i$ in the world coordinate system, respectively; $v_{k}^{i}$ is the speed of the $i$-th vehicle; $\psi_{k}^{i}$ denotes the heading angle of the $i$-th vehicle. 

\textbf{Action space $\mathcal{A}$}: In this work, a multi-discrete action space including three discrete sub-action spaces is adopted for the RL agent:
\begin{equation}
\begin{split}
\mathcal{A}=\left\{ A_{1},A_{2},A_{3} \right\},
\end{split}
\label{Action_space}
\end{equation}
where $A_{1}, A_{2}$, and $A_{3}$ represent the waypoint sub-action space, reference velocity sub-action space, and lane change sub-action space, respectively. 
The RL agent selects a set of actions within the $\mathcal{A}$. 
Then these actions are decoded to parameterize the low-level MPC to control the EV. 
Details will be illustrated in Section \ref{Section3-PPO}.

\textbf{State transition dynamics $\mathcal{P}(\mathbf{S}_{k+1}|\mathbf{S}_{k},a_{k})$}: It specifies the changes of environmental state, adhering the Markov property. It is implicitly defined by the external environment and cannot be accessed by the RL agent. 

\textbf{Reward function $\mathcal{R}$}: Here, we assign a positive reward if the RL agent finishes a target task to facilitate the improvement of the RL policy. A negative reward would be assigned if the agent maintains survival during the task process to incentivize the RL agent to finish the tasks efficiently. It also penalizes collisions and frequent lane-changing behaviors. Details will be introduced in Section \ref{Section3-PPO}.

\textbf{Discount factor $\gamma$}: $\gamma \in (0,1)$ is adopted for accumulated discount rewards in the future.

\section{Methodology}

\subsection{Overview of the Proposed Framework}
The workflow of the proposed method is illustrated in Fig. \ref{frame:whole}. 
Specifically, a novel Exp3.S-based BiMAB module is utilized to select the curricula, which determines the initial settings for the training environment of the RL agent. 
Then the RL agent selects actions based on its observation, which is then decoded and inputted into the MPC as high-level decision variables for parameterization. 
Subsequently, the parameterized MPC generates the control inputs based on the defined objective function and constraints, which are applied to the EV. Relevant information is recorded in the rollout buffer for updating the BiMAB policy and the RL policy. Details will be discussed in the following sections.

\subsection{Exp3.S-based BiLevel Multi-Armed Bandit Algorithm for Curriculum Learning}
\label{Section3-BiMAB}

Autonomous driving tasks at unsignalized intersections are typical instances of tasks with clustered structures. 
The number of SVs and the type of crossing task are the two main factors affecting the complexity of self-driving tasks at unsignalized intersections. 
The difficulty level increases rapidly as the number of SVs grows. This is because the EV needs to interact with SVs coming from various directions simultaneously, and these SVs have different destinations and driving intentions, introducing significant uncertainty into the tasks. 

\begin{figure}[!htbp] 
\centering
    \subfigure[SV from the upper region (L:2; S:1; R:1).]{\includegraphics[width=.15\textwidth]{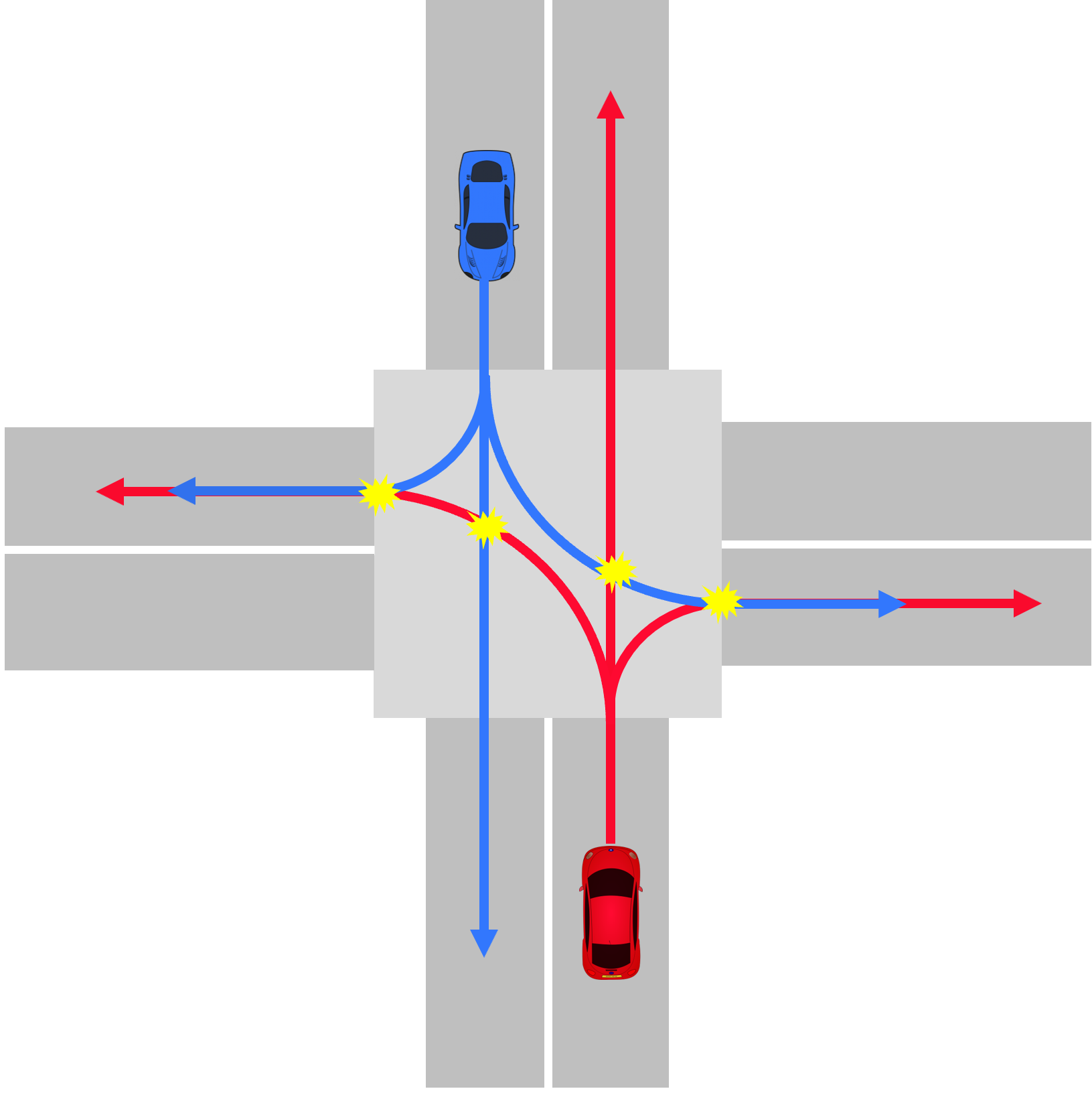}\label{cp:up}}
    \subfigure[SV from the left region (L:2; S:2; R:1).]{\includegraphics[width=.15\textwidth]{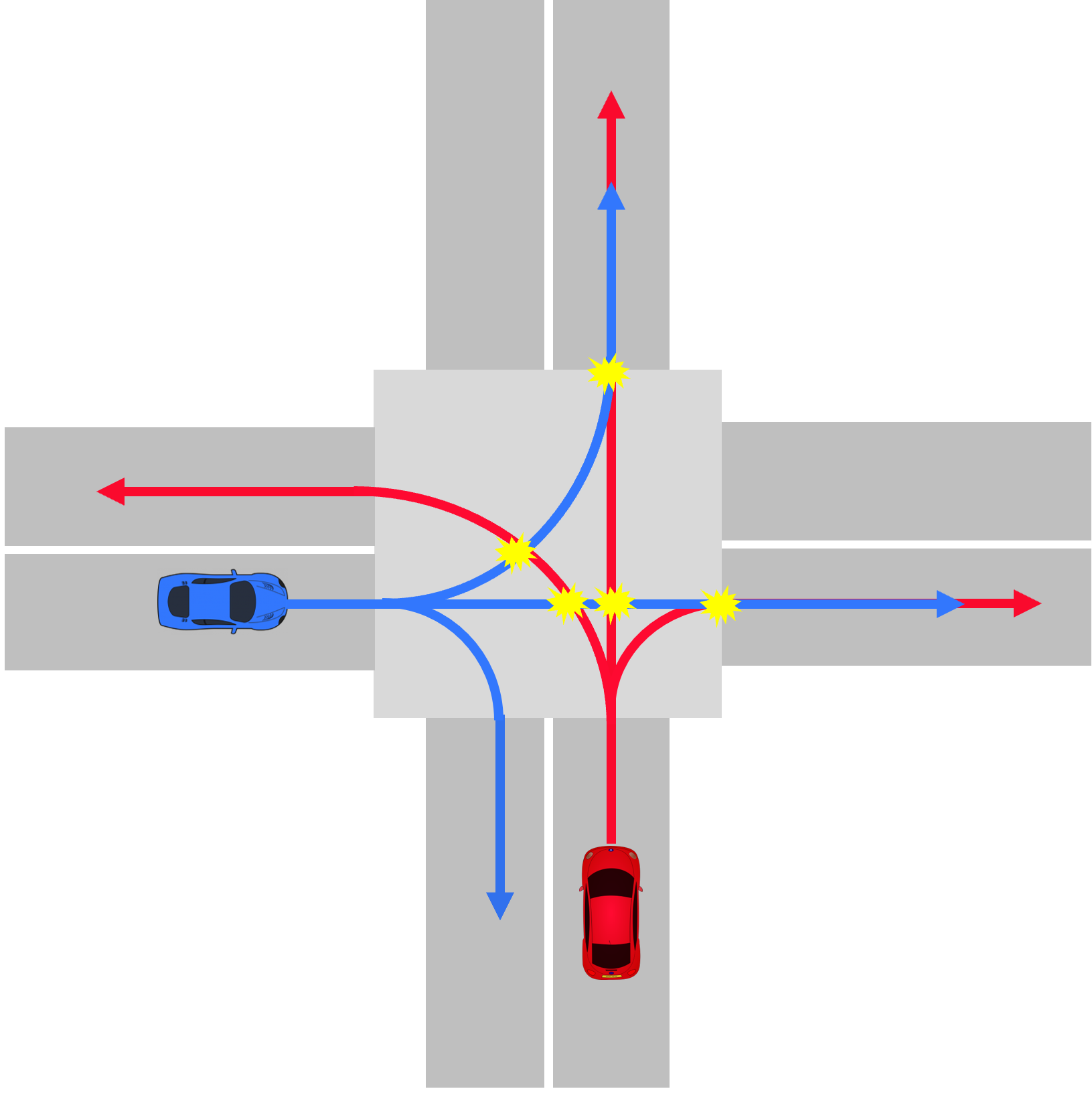}\label{cp:l}}
    \subfigure[SV from the right region (L:2; S:3; R:0).]{\includegraphics[width=.15\textwidth]{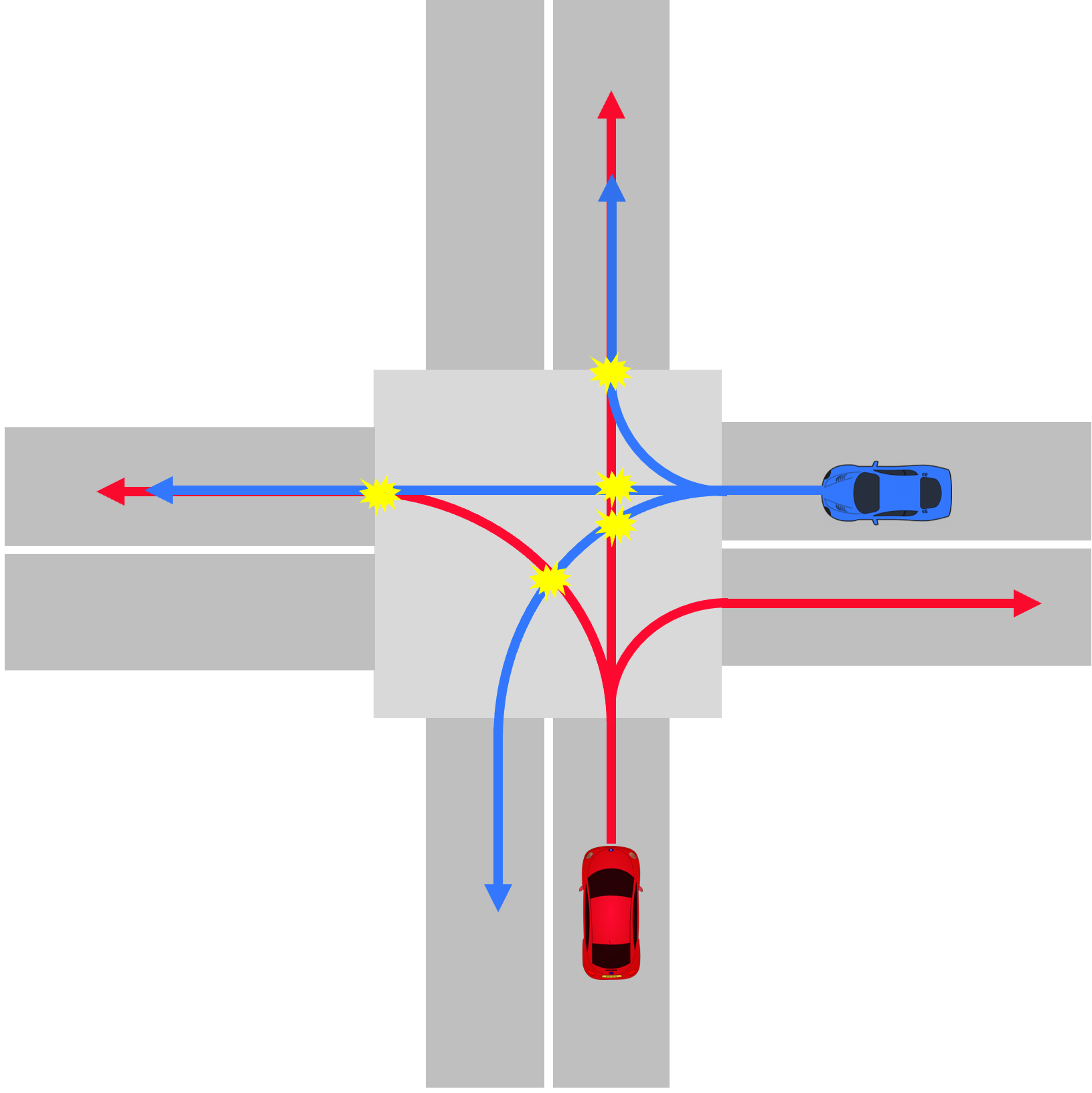}\label{cp:r}}
\caption{Potential collision points for EV (red car) crossing the intersection with the SV (blue car). L, S, and R represent the number of potential collision points when the EV performs diverse tasks.
}
\label{framework:coll_point}
\end{figure}

Different types of traversal tasks have various levels of challenge. 
The right-turn task, which involves completing a small radius turn along the road boundary and primarily dealing with the integration of other SVs into the target lane, is deemed the simplest. 
The go-straight task requires the EV to traverse through the central region of the intersection and potentially interact with SVs from three different directions, which is more complex than the right-turn task. 
The left-turn task, in addition to the challenges present in the go-straight task, necessitates a change in direction and experiences a longer distance in the central region, thereby rendering it the most difficult of the intersection driving tasks. 
A collision point analysis for left-turn, go-straight, and right-turn tasks at a single-lane intersection is shown in Fig. \ref{framework:coll_point}. The figure enumerates the number of collision points for each task when there is one SV in the scenario. As the number of SVs and lanes increases, the complexity of collision situations escalates, leading to a geometric growth in the task difficulty. 

The proposed BiMAB module is shown in Fig. \ref{frame:whole}, which is used for the allocation of training curricula.

\subsubsection{Task Decomposition and Bilevel Curriculum Modelling}

Based on the above analysis, we model the training process of the RL policy at unsignalized intersections as a bilevel curriculum learning task. 
Our curriculum set includes $N_{\text{ss}}^{\max}$ subsets, which are characterized by a progressively increasing number of SVs. Each subset consists of $N_{\text{sc}}^{\max}$ sub-curricula, representing different task types of EV within the corresponding subset. 
Specifically, the bilevel curriculum set is expressed as follows: 
\begin{equation}
\begin{split}
\mathbf{\Omega}=\left\{\Omega_{ij} | i = 0, 1, ..., N_{\text{ss}}^{\max}, j = 0, 1, ..., N_{\text{sc}}^{\max} \right\},
\end{split}
\label{Curr_Set}
\end{equation}
where $i$ denotes the serial number of the subset, which is equivalent to the number of SVs within corresponding curricula; $j$ is the serial number of the sub-curriculum, which means the type of the driving task, and $j=0,1,2$ represents a left-turn task, go-straight task, and right-turn task, respectively. 
It is noted that $N_{\text{sv}}^{\max}$ is equal to $N_{\text{ss}}^{\max}$. 
Within this set, each subset can be considered as a cluster, and correspondingly, each sub-curriculum can be regarded as an arm within the context of the BiMAB problem 
\cite{slivkins2019introduction,carlsson2021thompson}. 
Following the above analysis, the curriculum selection in (\ref{Curr_Set}) can be considered as a sampling process within the BiMAB $\mathcal{M}$, which consists of $N_{\text{ss}}^{\max}+1$ clusters and $N_{\text{sc}}^{\max}+1$ arms. 
Therefore, the BiMAB agent can derive a sample sequence of clusters and arms over $T$ episodes for the training of the RL agent:
\begin{equation}
 (C_T,A_T) = \left\{ (c_{s,1},a_{s,1}), (c_{s,2},a_{s,2}), ..., (c_{s,T},a_{s,T}) \right\},
\label{MAB_arm_sample}
\end{equation}
where $C_T$ and $A_T$ are the sampled cluster sequence and arm sequence across the RL training process, respectively; $c_{s,i}$ and $a_{s,i}$ are the sampled cluster and arm within the BiMAB in the $i$-th episode. 
When an episode ends, the rollout buffer of the RL agent returns a final reward. Then the BiMAB agent adjusts the importance weights based on this reward in conjunction with the historical rewards. 
Our objective is to develop an adaptive strategy to maximize payoffs generated from the sampled cluster and arm sequence $(C_T,A_T)$:
\begin{equation}
\mathbf{w}^{C*},\mathbf{w}^{A*}=\arg \max _{\mathbf{w}^C,\mathbf{w}^A,(C_T,A_T)} \sum\nolimits_{t=1}^T \hat{r}(t),
\label{Equ:MAB_max}
\end{equation}
where $\mathbf{w}^C$ and $\mathbf{w}^A$ are non-negative importance weight vectors of the clusters and arms in the BiMAB; $\mathbf{w}^{C*}$ and $\mathbf{w}^{A*}$ denote the optimal importance weight vectors of the clusters and arms in the BiMAB; $\hat{r}(t)$ represents the rescaled reward of the BiMAB, which depends on the reward received from the RL agent.

\subsubsection{Exp3.S-based BiMAB Algorithm for Automated Curriculum Selection} 
For the BiMAB model $\mathcal{M}$ constructed before, the importance weights $\mathbf{w}^C(t),\mathbf{w}^A(t)$ and the probability distributions $\mathbf{p}^C(t),\mathbf{p}^A(t)$ are defined as follows:
\begin{equation}
\begin{aligned}
\mathbf{w}^C(t)&=\left\{\mathrm{w}^c_i(t) | i = 0, 1, ..., N_{\text{ss}}^{\max} \right\},\\
\mathbf{p}^C(t)&=\left\{ p^c_i(t) | i = 0, 1, ..., N_{\text{ss}}^{\max} \right\},\\
\mathbf{w}^A(t)&=\left\{\mathrm{w}^a_{ij}(t) | i = 0, 1, ..., N_{\text{ss}}^{\max}, j = 0, 1, ..., N_{\text{sc}}^{\max} \right\},\\
\mathbf{p}^A(t)&=\left\{ p^a_{ij}(t) | i = 0, 1, ..., N_{\text{ss}}^{\max}, j = 0, 1, ..., N_{\text{sc}}^{\max} \right\}.
\end{aligned}
\label{def:weight_pobs}
\end{equation}

In our specific problem, the optimal cluster and optimal arm change across various stages of the training process due to the variation in the anticipated rewards, which are related to task settings and the improvement of the RL policy during the training process. 
Drawing inspiration from the Exp3.S algorithm \cite{auer2002nonstochastic}, we can tackle this problem by integrating an additional $\varepsilon$-greedy factor into the probability update mechanism. This ensures that each cluster and arm maintains a non-zero probability of selection throughout the training duration. 
With the importance weights $\mathbf{w}^C(t)$ and $\mathbf{w}^A(t)$ determined at the $t$-th episode, the sampling probabilities of the $i$-th cluster and $j$-th arm within the $i$-th cluster can be calculated as follows:
\begin{equation}
\begin{aligned}
p^c_i(t)=(1-\eta) \frac{e^{\mathrm{w}^c_i(t)}}{\sum_{k=0}^{N_{\text{ss}}^{\max}} e^{\mathrm{w}^c_k(t)}}+\frac{\eta}{N_{\text{ss}}^{\max}+1},\\ 
i=0, 1, ..., N_{\text{ss}}^{\max}, \\
p^a_{ij}(t)=(1-\eta) \frac{e^{\mathrm{w}^a_{ij}(t)}}{\sum_{k=0}^{N_{\text{sc}}^{\max}} e^{\mathrm{w}^a_{ik}(t)}}+\frac{\eta}{N_{\text{sc}}^{\max}+1},\\ 
i=0, 1, ..., N_{\text{ss}}^{\max}, j=0, 1, ..., N_{\text{sc}}^{\max},
\end{aligned}
\label{equ:pobs_exp_cal}
\end{equation}
where $\eta$ is a positive constant that balances the exploitation of obtained experiences and random exploration. 
Within the framework where the task exhibits a clustered structure, the BiMAB model introduces an equivalent update mechanism for the clusters. 
We first derive a cluster sample $c(t)$ at $t$-th episode according to the distribution $\mathbf{p}^C(t)$ as follows:
\begin{equation}
    c^\mathcal{M}(t) \sim \mathcal{M}(\mathbf{p}^C(t),\mathbf{p}^A(t)).
\label{equ:MAB_sample_cluster}
\end{equation}

Then the BiMAB agent chooses an arm sample $a^c(t)$ for the RL agent from the sampled cluster $c^\mathcal{M}(t)$ according to corresponding possibility distribution $\mathbf{p}^A(t)$ as follows: 
\begin{equation}
    a^c(t) \sim \mathcal{M}(\mathbf{p}^A(t)|c^\mathcal{M}(t)).
\label{equ:MAB_sample_arm}
\end{equation}

Once the cluster and arm are sampled, the RL environment is initialized according to the corresponding curriculum settings. 
The BiMAB agent will obtain a reward $r_{ij}(t)$ from the RL agent when this episode ends. 
After the RL agent finishes an episode, receiving a positive reward indicates a successful interaction with the environment that meets expectations, whereas receiving a negative reward suggests the occurrence of undesired events or dangerous situations such as collisions during that episode. For the RL policy, negative rewards mean poor performance in that scenario, indicating the need for further training. Based on the above analysis, it is essential for the BiMAB module to appropriately modulate the received negative rewards to facilitate the training of the RL policy. 
In this work, the rescaled reward for the BiMAB is defined as follows:
\begin{equation}
\left\{
\begin{array}{l}
\hat{r}_{i}(t)=\frac{r^{ij}_{\text{norm}}(t)}{p_i(t)}, \\
\hat{r}_{ij}(t)=\frac{r^{ij}_{\text{norm}}(t)}{p_{ij}(t)}, \\
r^{ij}_{\text{norm}}(t)=\frac{2\left(r_{\text{md}} - k_0 R_{\min}(t)\right)}{k_1 R_{\max}(t) - k_0 R_{\min}(t)}-1, 
\end{array}
\right.
\label{Exp3:rescale}
\end{equation}
where
\begin{equation}
r_{\text{md}} = \left\{
\begin{array}{l}
r_{ij}(t), \ \text{if}\ \ r_{ij}(t) \geq 0, \\
-\alpha_{\text{md}} r_{ij}(t), \  \text{otherwise},
\end{array}
\right.
\end{equation}
where $\hat{r}_i(t)$ and $\hat{r}_{ij}(t)$ represent the rescaled reward for $i$-th cluster and corresponding $j$-th arm within the $i$-th cluster obtained in $t$-th episode, respectively; 
$R_{\max}(t)$ and $R_{\min}(t)$ are the maximum and minimum absolute values of the unscaled rewards in the history up to the current episode, respectively; 
$k_0$, $k_1$, and $\alpha_{\text{md}}$ denote positive constants for the rescaling procedure. 
The reward for both cluster and arm that are not sampled in the $t$-th episode is 0. 
Here, we perform an absolute value operation on the received reward value, aiming to enable the BiMAB agent to utilize information related to the courses where the RL policy is currently underperforming. 
Additionally, we proportionally scale the normalized reward $r^{ij}_{\text{norm}}(t)$ for the corresponding cluster and arm based on their selection probability. This allows clusters and arms that could be optimal but currently have low probabilities to be timely identified through algorithm updates. 
Then the importance weights of $i$-th cluster, and $j$-th arm within $i$-th cluster are adjusted as follows:
\begin{equation}
\begin{aligned}
\mathrm{w}^c_i(t+1)&=\mathrm{w}^c_i(t) + \alpha_c \hat{r}_i(t) + \beta_c \mathrm{W}_{i}(t),\\
\mathrm{w}^a_{ij}(t+1)&=\mathrm{w}^a_{ij}(t) + \alpha_a \hat{r}_{ij}(t) + \beta_a \mathrm{W}_{ij}(t),
\end{aligned}
\label{equ:weight_update}
\end{equation}
where $\mathrm{W}_{i}(t)=\sum_{i=0}^{N^{\max}_{\text{ss}}} \mathrm{w}^c_i(t)$, $\mathrm{W}_{ij}(t)=\sum_{j=0}^{N^{\max}_{\text{sc}}} \mathrm{w}^a_{ij}(t)$; $\alpha_c$, $\alpha_a$, $\beta_c$, and $\beta_a$ are positive constant parameters for adjusting the growth rate of importance weights of each cluster and each arm, respectively.

Based on the calculated importance weights, specific curricula are sampled for training the RL policy. 
In practical deployment, due to the inherent randomness of sampling and the stochastic generation of SVs, it is difficult to avoid inefficient samples completely during the training process. 
To prevent occasional sampling that leads to misestimation towards the importance of clusters and arms, we utilize a target BiMAB $\hat{\mathcal{M}}$ to stabilize the update process of BiMAB $\mathcal{M}$. 
After receiving the final reward from the RL agent, the target BiMAB agent updates its parameters based on the rescaled rewards. Then, after a certain number of curricula are sampled, the importance weights of the target BiMAB are synchronized to the BiMAB to proceed with the selection of curricula for the next phase. 
Therefore, the update of importance weights within the BiMAB can be rewritten as follows:
\begin{equation}
\begin{aligned}
\mathrm{w}^c_i(t+1)&=\left\{
\begin{array}{l}
\mathrm{w}^c_i(t),\ \text{if}\ \ t | N_{BiMAB} \neq 0, \\
\mathrm{w}^c_i(t) + \sum_{k=t-N_{BiMAB}+1}^{t} (\alpha_c \hat{r}_i(k) \\
\qquad\qquad\qquad\qquad + \beta_c \mathrm{W}_{i}(k)) ,\ \text{otherwise},
\end{array}
\right. \\
\mathrm{w}^a_{ij}(t+1)&=\left\{
\begin{array}{l}
\mathrm{w}^a_{ij}(t),\ \text{if}\ \ t | N_{BiMAB} \neq 0, \\
\mathrm{w}^a_{ij}(t) + \sum_{k=t-N_{BiMAB}+1}^{t} (\alpha_a \hat{r}_i(k) \\
\qquad\qquad\qquad\qquad + \beta_a \mathrm{W}_{ij}(k)) ,\ \text{otherwise},
\end{array}
\right.
\end{aligned}
\label{equ:weight_double}
\end{equation}
where $N_{BiMAB}$ is the update interval for synchronization between the target BiMAB $\hat{\mathcal{M}}$ and the BiMAB $\mathcal{M}$.

The complete procedure of the proposed Exp3-based BiMAB algorithm for automated curriculum selection is summarized in \textbf{Algorithm \ref{ALG_BiMAB}}.

\begin{algorithm}[!htbp] 
	\caption{Exp3.S-based BiMAB Algorithm} 
        \label{ALG_BiMAB}
	\LinesNumbered 
	\KwIn{Curriculum set $\mathbf{\Omega}$, BiMAB update frequency $N_{BiMAB}$}
	\KwOut{Selected cluster $c^\mathcal{M}$, selected arm $a^c$}
        Initialize the BiMAB and target BiMAB with weights $\left\{ \mathrm{w}_i^c \right\}$, $\left\{ \hat{\mathrm{w}}_i^c \right\}$, $\left\{ \mathrm{w}_{ij}^a \right\}$, and $\left\{ \hat{\mathrm{w}}_{ij}^a \right\}$, where $i=0,1,...,N_{\text{ss}}^{\max},j= 0, 1, ..., N_{\text{sc}}^{\max}$\; 
        \While{$t \leq t_{\max}$}  
            {Compute probability distributions of clusters $\mathbf{p}_c(t)$ from (\ref{equ:pobs_exp_cal})\;
            Derive a sample cluster $c^\mathcal{M}(t)$ from (\ref{equ:MAB_sample_cluster})\;
            Compute probability distributions of arms $\mathbf{p}^C_a(t)$ in cluster $c^\mathcal{M}(t)$ from (\ref{equ:pobs_exp_cal})\;
            Derive a sample arm $a^c(t)$ from (\ref{equ:MAB_sample_arm})\;
            Play arm $a^c(t)$ and apply corresponding $\Omega(t)$ to the RL environment for initialization\;
            Receive the terminal reward $r_{ij}(t)$ from the RL agent\;
            Compute the rescaled reward $\hat{r}_i(t)$ and $\hat{r}_{ij}(t)$ from (\ref{Exp3:rescale})\;
            Update the corresponding cluster and arm weight within the target BiMAB from (\ref{equ:weight_update})\;
            \If{$t\ |\ N_{BiMAB}=0\ $}
                {$\mathrm{w}_i(t)=\hat{\mathrm{w}}_i(t),\ i=0,1,...,N_{\text{ss}}^{\max}$\;
                $\mathrm{w}_{ij}(t)=\hat{\mathrm{w}}_{ij}(t),\ i=0,1,...,N_{\text{ss}}^{\max},\ j=0,1,...,N_{\text{sc}}^{\max}$\;}
            }
\end{algorithm}

\subsection{High-Level PPO for Decision-Making}
\label{Section3-PPO}

Within the proposed framework, the RL policy is employed to generate high-level decision variables, which account for the trade-off between safety and efficiency throughout the autonomous driving task process. 
Specifically, the RL agent aims to decide a sequence of the intermediate points as the reference of low-level MPC. We use the notations of subscript $k$ to represent  $k$-th time step within an episode. 

\subsubsection{Observation and Action Generation}

Here, we define the observations for the RL agent as follows:
\begin{equation}
\begin{split}
\mathbf{O}_{k}=\left[\ \mathbf{o}_{k}^0\ \ \mathbf{o}_{k}^1\ ...\ \mathbf{o}_{k}^{N_{\text{sv}}^{\max}}\ \right]^T,
\end{split}
\label{state martix}
\end{equation}
where $\mathbf{o}_{k}^0 = [\tilde{x}_k^0\ \tilde{x}_k^0\ v^0_k\ \tilde{\psi}^0_k]^T$; $\tilde{x}_k^0$ and $\tilde{y}_k^0$ represent the absolute differences between the current X-axis and Y-axis coordinates of the EV and these of the goal location, respectively; $v^0_k$ denotes the current speed of the EV; $\tilde{\psi}^0_k$ is the difference between the current heading angle and the goal heading angle. $\mathbf{o}_{k}^i\ (i=1,2,...,N^{\max}_{\text{ss}})$ is defined as follows:
\begin{equation}
\begin{split}
\mathbf{o}_{k}={\left[\begin{array}{l l l l l l l}{dx_{k}^{i}}&{dy_{k}^{i}}&{dv_{k}^{i}}&{d\psi_{k}^{i}}\end{array}\right]}^{T},
\end{split}
\label{observation}
\end{equation}
where
\begin{equation}
\left\{\begin{array}{l}
dx_{k}^{i} = \textup{clip}(|x_{k}^{0}-x_{k}^{i}|,0,7.5), \\
dy_{k}^{i} = \textup{clip}(|y_{k}^{0}-y_{k}^{i}|,0,7.5), \\
dv_{k}^{i} = v_{k}^{0}-v_{k}^{i}, \\
d\psi_{k}^{i} = \psi_{k}^{0}-\psi_{k}^{i}, \\
\end{array}\right.
\end{equation}
where $\textup{clip}(\cdot)$ is a clip function to pre-process the corresponding elements to constrain them within a predefined range. 

In this work, the RL policy is represented by a neural network $\pi$ parameterized by $\bm{\theta}$. Given the RL observation $\mathbf{O}_k$ at time step $k$, the action of RL agent is generated by:
\begin{equation}
    a_k^{RL} = \pi_{\bm{\theta}}(\mathbf{O}_k)
    \label{eq:get_action}
\end{equation}

\begin{figure}[!htbp] 
\centering
    \subfigure[Update process of $A_1$.]{\includegraphics[width=.22\textwidth]{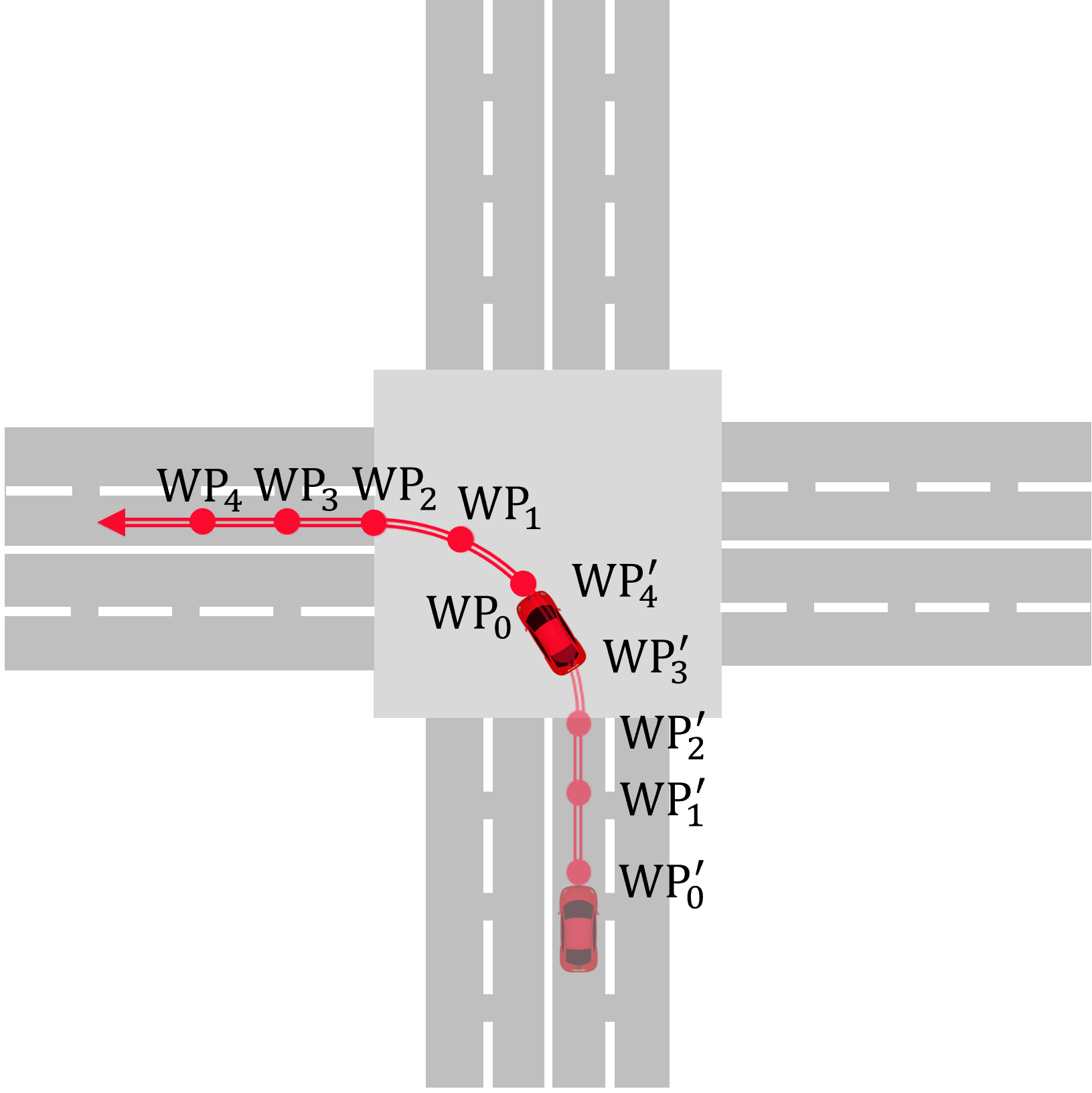}\label{wp_a1_1_2}}
    \subfigure[Lane-changing behavior.]{\includegraphics[width=.22\textwidth]{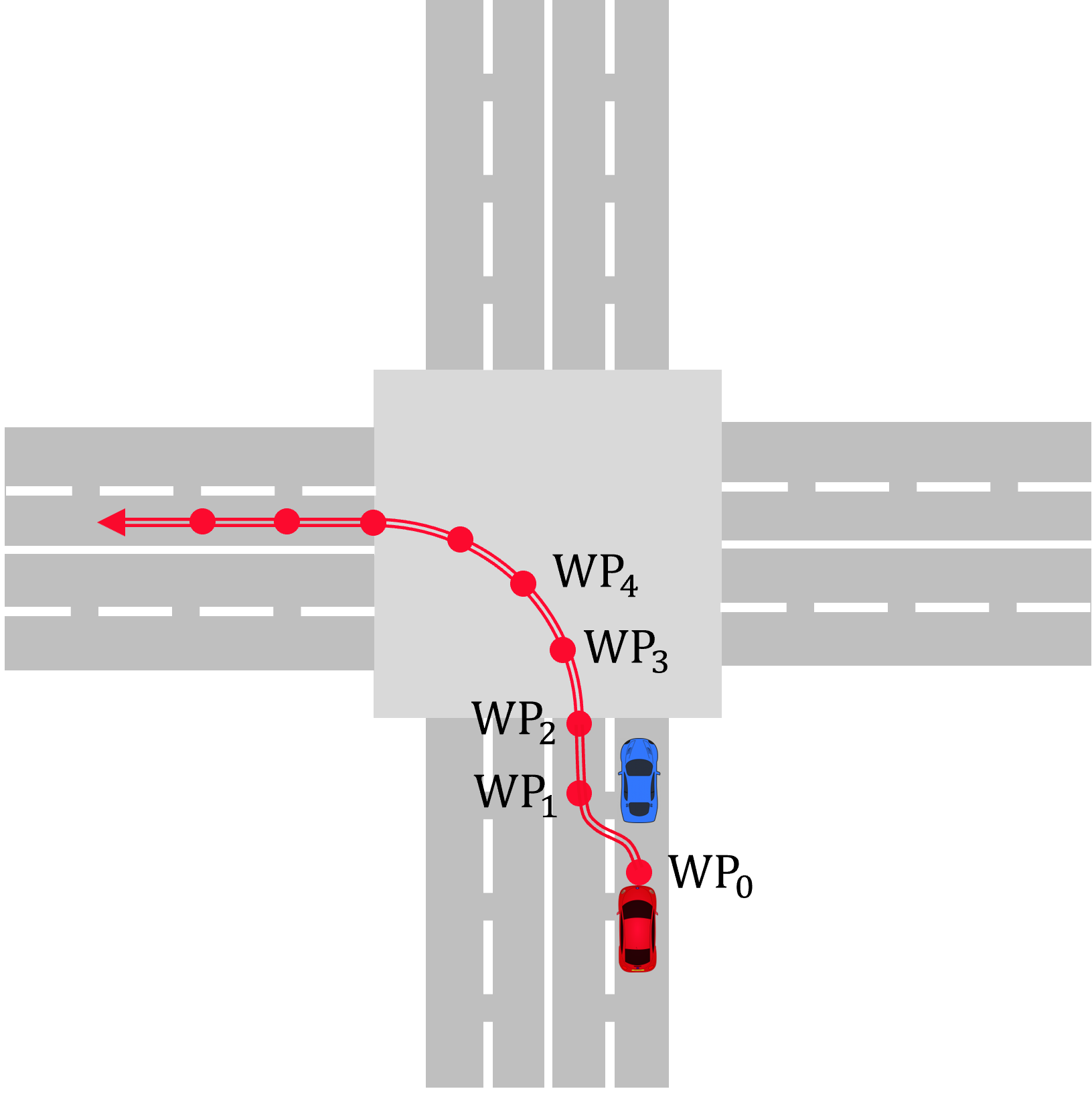}\label{wp_a3}}
\caption{Visualization of designed action space of RL agent. 
}
\label{framework:waypoint}
\end{figure}

The specific definition of the sub-action spaces in (\ref{Action_space}) are introduced as follows. 
The waypoint sub-action space is defined as:
\begin{equation}
    A_1 = \left\{ \textup{WP}_0, \textup{WP}_1, ..., \textup{WP}_4\right\}, 
    \label{AS1}
\end{equation}
where $\textup{WP}_i$ in $A_1$ represents the $i$-th waypoint, which can be represented as follows:
\begin{equation}
    \textup{WP}_i = [x^{\text{WP}}_i\ y^{\text{WP}}_i\ \psi^{\text{WP}}_i]^T,
\end{equation}
where $x^{\text{WP}}_i,y^{\text{WP}}_i$, and $\psi^{\text{WP}}_i$ denote the reference information about the X-axis, Y-axis coordinates, and heading angle of the waypoint, respectively. 
Waypoints are provided by a predefined road map and several path-searching methods, such as $A^*$ search algorithm. 
The visualization of the sub-action space $A_1$ is shown in Fig. \ref{wp_a1_1_2}, where the EV drives from the lower region towards the central zone. A reference waypoint set is generated using A* search at the beginning of the task. When the EV is in the lower area, the 5 waypoints closest to the EV (${\textup{WP}^{\prime}_{i},i=0,1,...,4}$) are added to the $A_1$. 
Before each decision-making process, the RL agent checks the relative positions of all waypoints in the reference waypoint set against the position of the EV, and updates the sub-action space $A_1$. Specifically, any waypoints located behind the current position of the EV are removed from the reference waypoint set. 
Then the 5 waypoints closest to the EV (${\textup{WP}_{i},i=0,1,...,4}$) in the new updated reference waypoint set comprise $A_1$. 
The RL agent then selects the appropriate action from this updated sub-action space $A_1$. It is worth noting that both the reference waypoint set and $A_1$ are consistently checked and updated throughout the task. 
Specifically, the first waypoint $\textup{WP}_0$ is the waypoint closest to the EV, thereby enabling the functionality for stopping and yielding. 
Distant waypoints guide the EV towards the target position in diverse modes. 
According to \cite{speedlimit_fhwa,patil2016microscopic}, the maximum reference speed is set to 8 m/s. 
The reference velocity sub-action space is defined as:
\begin{equation}
    A_2 = \left\{ 0,2,4,6,8\right\}, 
    \label{AS2}
\end{equation}
where the unit of the elements is m/s. The lane change sub-action space is defined as:
\begin{equation}
    A_3 = \left\{ -1,0,1 \right\},
    \label{AS3}
\end{equation}
where $-1,0$, and $1$ represent left lane change, lane keeping, and right lane change maneuvers, respectively. The visualization of $A_3$ can be referred to in Fig. \ref{wp_a3}. In Fig. \ref{wp_a3}, the EV detects an SV ahead, so it changes lanes to the left to avoid a potential collision and prepare for the left-turn maneuver.

By configuring the action space in this manner, the RL agent is not only equipped with navigation and obstacle avoidance capabilities but also can be generalized to various driving scenarios that feature regular road structures and waypoints. 

\begin{remark}
    By constructing an action space composed of (\ref{AS1}), (\ref{AS2}), and (\ref{AS3}), we can facilitate flexible motion patterns for the EV to interact with SVs exhibiting multi-modal behaviors. 
    When the EV needs to drive towards the target location rapidly, the RL agent can select the most distant waypoint and a high reference speed. Conversely, when the EV requires an emergency brake to yield, the RL agent can choose the nearest waypoint and a low reference speed. Moreover, the introduction of a lane-changing sub-action space further enhances the flexibility of motion for the EV. When the RL agent decides to change lanes to the left or right, the selected waypoint will be replaced by the corresponding waypoint on the adjacent lane. Choosing a closer waypoint on the adjacent lane indicates an urgent lane change to avoid collisions, while picking a further waypoint on the adjacent lane means a smooth lane change for collision avoidance or overtaking.
\end{remark}

\subsubsection{Reward Function}

The reward function is crucial for motivating RL agents to explore the environment. 
Therefore, it is essential to design an appropriate reward function that is able to guide the RL agent to balance driving safety and efficiency during the training process. Taking into account the specific characteristics of the target scenarios addressed in this study, the reward function is designed as follows:
\begin{equation}
\begin{split}
    r(k) &=  r_{\text{S}}(I_{\text{S}}) + r_{\text{F}}(\neg I_{\text{S}}) + r_{\text{C}}(I_{\text{C}}) + r_{\text{TO}}(I_{\text{TO}})\\ 
    & \quad + r_{\text{LC}}(I_{\text{LC}}) + r_{\text{l}},
\end{split}
\label{reward_func}
\end{equation}
where $r_{\text{S}}$ and $r_{\text{F}}$ denote the rewards for task completion and failure by the RL agent, respectively; $r_{\text{C}}$, $r_{\text{TO}}$, $r_{\text{LC}}$, and $r_{\text{l}}$ denote the penalty for collisions, time-out situations, lane-changing behaviors, and surviving in the task, respectively. $I_{event}$ serves as an indicator function, marking the occurrence of diverse events. It is defined as follows:
\begin{equation}
I_{event}=\left\{
\begin{array}{l}
1,\ \text{if}\ event\ \text{occurs}, \\
0,\ \text{otherwise},
\end{array}
\right.
\end{equation}

To facilitate curriculum switching, the success reward term $r_{\text{S}}$ is configured to depend on the number of SVs and potential collision points, which can encourage the RL agent to explore the tasks with high difficulty levels. 
To assist the RL agent in learning from failed episodes, we set the failure reward $r_{\text{F}}$ to be related to the distance $d^{\text{F}}_{\text{e2g}}$ of the EV from the goal location when the episode ends. 
Furthermore, to improve the interaction awareness of the RL agent, the penalty for collisions $r_{\text{C}}$ is structured to have a positive correlation with both the speed of the EV $v_{\text{EV}}$ and the count of SVs. Furthermore, the penalty for lane-changing maneuvers is added to avoid non-compliant and frequent lane-changing behaviors. 
Among these reward terms, the success reward $r_{\text{S}}$, the failure reward $r_{\text{F}}$, and the collision penalty $r_{\text{C}}$ are designed as follows:
\begin{equation}
\begin{aligned}
r_{\text{S}}(I_{\text{S}}) &= I_{\text{S}} \cdot (\alpha_1 \cdot N_{\text{sv}} + \alpha_2 \cdot N_{\text{pcp}}),\\
r_{\text{F}}(\neg I_{\text{S}}) &= \min (r_{\text{F},\max},\neg I_{\text{S}} \cdot \alpha_3 \cdot {d^{\text{F}}_{\text{e2g}}}^{-1}),\\
r_{\text{C}}(I_{\text{C}}) &= I_{\text{C}} \cdot \alpha_4 \cdot N_{\text{sv}} \cdot v_{\text{EV}},
\end{aligned}
\label{def:weight_pobs}
\end{equation}
where $N_{\text{pcp}}$ denotes the number of potential collision points; $r_{\text{F},\max}$ is the maximum reward for the failed episodes; $\alpha_i$ are constant parameters, where $i=1,2,\cdots,4$. 
The remaining reward terms are assigned constant values.

\subsubsection{Training with BiM-ACPPO}

After sampling a cluster and a corresponding arm in the BiMAB, the RL agent explores the environment set by the configuration of the selected curriculum. 
Relevant observations, actions, and rewards within respective episodes are recorded in the replay buffer. 
Once a certain number of episodes have been gathered, the RL policy undergoes training to optimize the cumulative objective function that is associated with the sequence of sampled cluster-arm pairs $(C_T,A_T)$ as follows:
\begin{equation}
\bm{\theta}^*=\arg \max _{\bm{\theta},(C_T,A_T)} J(\bm{\theta}),
\label{equ:ACRL}
\end{equation}
where $J(\bm{\theta})$ denotes the objective function for the RL policy with parameter $\bm{\theta}$. 
Here, the clipped objective function of the PPO algorithm \cite{schulman2017proximal} is utilized for the update of the RL policy:
\begin{equation}
J_k(\bm{\theta})=\mathbb{E}_k\left[\min \left(\rho_k(\bm{\theta}) \hat{A}_k, \operatorname{clip}\left(\rho_k(\bm{\theta}), 1-\epsilon, 1+\epsilon\right) \hat{A}_k\right)\right],
\end{equation}
where $\rho_k(\bm{\theta})$ represents the probability ratio of the new policy to the old policy; $\hat{A}_k$ denotes the estimator of the advantage function at time step $k$; $\epsilon$ is the clip parameter. 
To reduce the variance of the long-term return estimation, the generalized advantage estimation (GAE) is adopted. Furthermore, to balance the exploration and exploitation, and also stabilize the policy update,  a temporary policy network is utilized for the implementation.

\subsection{Low-Level RL-Guided Model Predictive Control}
Upon receiving the current observation, the RL policy chooses high-level decision variables, which are then decoded and passed into the low-level MPC to generate control actions. 

The MPC is utilized for the motion control of the EV at the low level of the proposed framework. Typically, MPC uses the target location as the reference state in the receding horizons to force the EV to reach its destination swiftly. However, incorporating highly non-convex and nonlinear collision avoidance constraints into the MPC formulation is necessary to ensure the safety of the EV, which significantly increases the computational burden, especially in complex driving scenarios such as unsignalized intersections. In this work, we introduce an intermediate reference state to guide the EV to avoid collisions, which is determined by RL policy. 
This replaces the target location as the reference state of the MPC formulation. The intermediate reference state vector is defined as follows:
\begin{equation}
     \mathbf{x}_{\text{IMR}} =
    \begin{bmatrix} x_{\text{IMR}}& y_{\text{IMR}}& v_{\text{IMR}}& \psi_{\text{IMR}} \end{bmatrix} ^T 
    \label{eq:def_IMR}
\end{equation}
where $x_{\text{IMR}}$ and $ y_{\text{IMR}}$ are the intermediate reference X-axis and Y-axis coordinates of the EV in the world coordinate system, respectively; $v_{\text{IMR}}$ is the intermediate reference speed of the EV; $\psi_{\text{IMR}}$ denotes the intermediate reference heading angle of the EV. 
Here, we replace the goal point $\mathbf{x}_g$ in (\ref{mpc_org}) with the intermediate point which is determined by the RL agent. 

In the proposed framework, the MPC is tasked with tracking the intermediate reference states generated by the RL policy, while ensuring both the reduction of vehicle energy consumption and the maintenance of driving comfort.  
Given the high-level decision variable $\mathbf{x}_{\text{IMR}}$ and the current state of the EV $\mathbf{x}_k=[x_k\ y_k\ v_k\ \psi_k]^T$, the autonomous driving task can be formulated as an MPC problem over the receding horizon $N$. 
Considering the non-holonomic dynamics constraint and box constraints of the velocity and control commands of the EV, a nonlinear and nonconvex-constrained optimization problem can be formulated as follows:
\begin{equation}
\begin{aligned}
&\min _{\mathbf{x}_{1: N}, \mathbf{u}_{0: N-1}} J_{\mathbf{x}_{N}} + \sum_{k=0}^{N-1}\left(J_{\mathbf{x}_k}+ J_{\mathbf{u}_k}+J_{\Delta \mathbf{u}_k}\right)\\
&\text {\quad \quad s.t.} \quad \mathbf{x}_{k+1}=\mathbf{x}_k+f\left(\mathbf{x}_k, \mathbf{u}_k\right) d_k, \\
 & \quad \quad \quad \quad \mathbf{u}_{\min } \leq \mathbf{u}_k \leq \mathbf{u}_{\max }, \\ 
 & \quad \quad \quad \quad v_{\min } \leq v_k \leq v_{\max }, \\ 
 & \quad \quad \quad \quad \mathbf{x}_0=\mathbf{x}_{\text{current}},\\
\label{mpc_re} 
\end{aligned}
\end{equation}
where
\begin{equation}
    \begin{aligned}
        J_{\mathbf{x}_{N}} &= (\mathbf{x}_N - \mathbf{x}_{\text{IMR}})^T \mathbf{Q}_x(\mathbf{x}_N - \mathbf{x}_{\text{IMR}}),\\
        J_{\mathbf{x}_{k}} &= (\mathbf{x}_k - \mathbf{x}_{\text{IMR}})^T \mathbf{Q}_x(\mathbf{x}_k - \mathbf{x}_{\text{IMR}}),\\
        J_{\mathbf{u}_{k}} &= \mathbf{u}_k^T \mathbf{Q}_u \mathbf{u}_k,\\
        J_{\Delta \mathbf{u}_k} &= \Delta \mathbf{u}_k^T \mathbf{Q}_{\Delta u} \Delta \mathbf{u}_k,\\
    \end{aligned}
\end{equation}
where $\mathbf{\Delta u}_k = u_k - u_{k-1}$ is the variation of control commands; $\mathbf{x}_{\text{current}}$ represents the current state vector of the EV; $\mathbf{Q}_x, \mathbf{Q}_u$, and $\mathbf{Q}_{\Delta u}$ are positive semi-definite diagonal weighting matrices for corresponding terms.

The complete procedure of the proposed BiM-ACPPO framework is summarized in \textbf{Algorithm \ref{ALG_BiMAB_ACPPO}}.

\begin{algorithm}[t]  
	\caption{BiM-ACPPO} 
        \label{ALG_BiMAB_ACPPO}
	\LinesNumbered 
	\KwIn{Environmental state $\mathbf{S}_{k}$, 
        RL policy update frequency $N_{\text{RL}}$}
	\KwOut{$\pi_{\bm{\theta}^*} = f(\bm{\theta}^*)$}
        Initialize the policy network and the temporary policy network with parameter $\bm{\theta}_0$, $\bm{\theta}_0^t$\;
        Derive a sample $\Omega(t)$ according to Algorithm \ref{ALG_BiMAB}\;
        Initialize the RL environment based on the sampled curriculum $\Omega(t)$\;
        \While{$t \leq t_{\max}$}  
            {
            \While{not done}
            {RL agent chooses high-level decision variables from (\ref{eq:get_action})\;
            Decode the high-level decision variables to obtain the intermediate reference vector $\mathbf{x}_{\text{IMR}}$\; 
            Parameterize the MPC (\ref{mpc_re}) and calculate the control action\;
            Apply the control action to the EV, obtain the reward $r_k(t)$ and the next state $r_{k+1}(t)$;\
            }
            Record related information of episodes experienced by the RL agent, including the history of states, actions, and rewards\;
            Update the temporary policy network by (\ref{equ:ACRL})\;
            Feed the reward at the end of the $t$-th episode to BiMAB in Algorithm \ref{ALG_BiMAB}\;
            \If{$t\ |\ N_{\text{RL}}=0$}
                {Synchronize the parameters of the temporary policy network to the current policy network\;
                }
            Derive a sample $\Omega(t)$ according to Algorithm \ref{ALG_BiMAB}\;
		  Reset the unsignalized intersection based on the sampled curriculum $\Omega(t)$\;
            }
            Save the final policy network $\pi_{\bm{\theta}}^*=f(\bm{\theta}^*)$.\
\end{algorithm}
\vspace{-0.1cm}

\section{Experiments}

\subsection{Experimental Setup}

In this section, we implement the BiM-ACPPO approach in two different unsignalized intersections and an overtaking scenario. 
The experiments are carried out on the Ubuntu 18.04 system with 2.60GHz Intel(R) Xeon(R) Platinum 8358P CPU and NVIDIA GeForce RTX 4090 GPU. All self-driving scenarios involved in experiments are constructed on the CARLA simulator \cite{dosovitskiy2017carla}. 
Here, we set the Tesla Model 3 as the self-driving vehicle. 
The actor-critic architecture is adopted to implement the proposed method. 
The action network and critic network are set as fully connected networks with 2 hidden layers of 256 units and 128 units by PyTorch and trained with the Adam optimizer. 
The number of epochs is set to 20. The learning rate of the action network and critic network are set to $5 \times 10^{-4}$ and $1 \times 10^{-3}$, respectively. $\gamma$ is set to 0.99. 
The initial weights of BiMAB are set to $\mathrm{w}^c_i(0)= 1,\mathrm{w}^a_{ij}(0),i =0,1,...,N_{\text{ss}}^{\max}, j = 0, 1, ..., N_{\text{sc}}^{\max}$. $\eta$ is set to 0.2. $N_{BiMAB}$ is set to 1000. 
The low-level MPC optimization problem is solved by CasADi \cite{andersson2019casadi}, with the IPOPT option and single-shooting approach. The weighting matrices $\mathbf{Q}_x$, $\mathbf{Q}_u$, and $\mathbf{Q}_{\Delta u}$ are set to $\operatorname{diag}\left(\left[100, 100, 100, 20\right]\right)$, $\operatorname{diag}\left(\left[10, 10\right]\right)$, and $\operatorname{diag}\left(\left[1, 1\right]\right)$, respectively. 
In this work, the proposed framework is compared with four baseline approaches as follows: 

\begin{itemize}
    \item Fixed PPO: the RL policy is directly trained by PPO approach \cite{schulman2017proximal} in the target scenario with $N_{\text{sv}}=N_{\text{sv}}^{\max}$.
    \item Random CPPO: the probability distributions of all clusters and arms in the curriculum set are equal.
    \item Manual CPPO: the RL policy is trained by a simplified implementation of \cite{peng2023CPPO} with a fixed $\epsilon$.
    \item RD-ACPPO: implementation of the approach in \cite{peng2024rewarddriven} with MPC as the low-level controller.
\end{itemize}

For the sake of fairness, we set the clip parameters of all methods to $\epsilon=0.2$. 
The SVs are in built-in autopilot mode which is provided by the CARLA simulator. 
We first train the RL policy in the unsignalized intersection scenario. 
Then the test of all trained RL policies at unsignalized intersections with different numbers of SVs $N_{\text{sv}}=0,1,...,N_{\text{sv}}^{\max}$ and diverse driving tasks are conducted. 

\vspace{-0.2cm}

\begin{figure*}[!htbp] 
\centering
    \subfigure[First cluster ($N_{\text{ss}}=0$)]{\includegraphics[width=.24\textwidth]{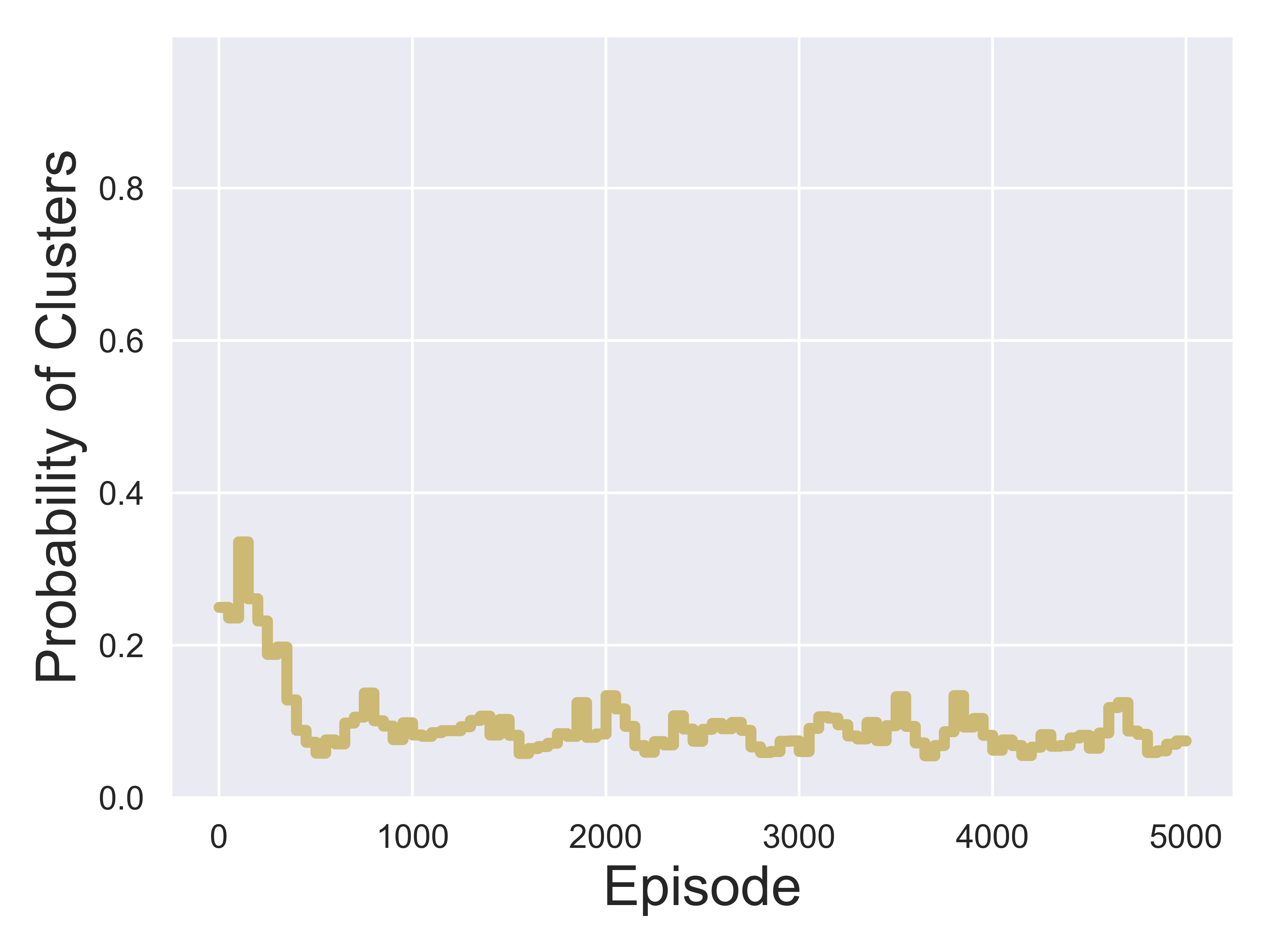}\label{fig_res_cls0}}
    \subfigure[Second cluster ($N_{\text{ss}}=1$)]{\includegraphics[width=.24\textwidth]{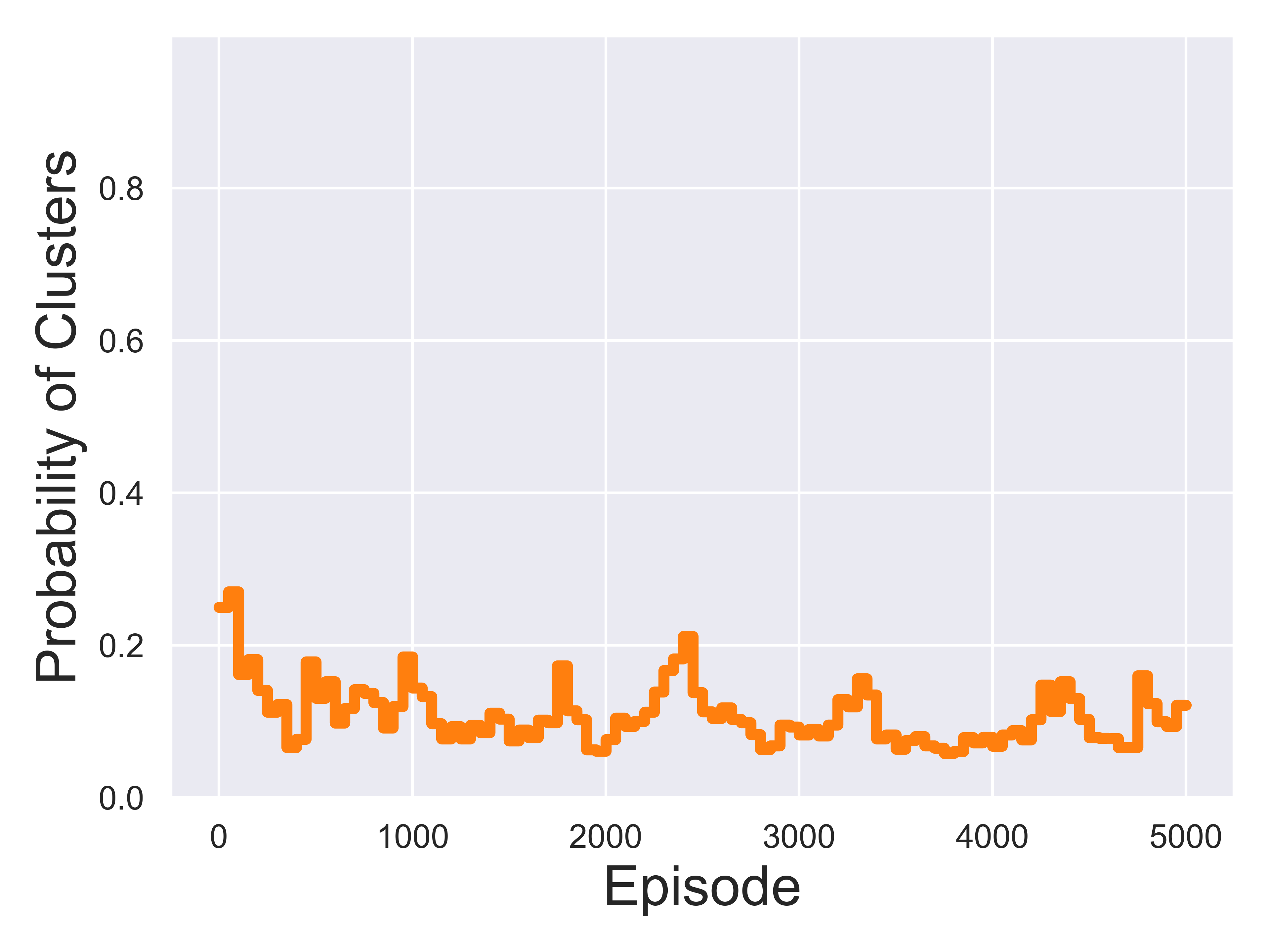}\label{fig_res_cls1}}
    \subfigure[Third cluster ($N_{\text{ss}}=2$)]{\includegraphics[width=.24\textwidth]{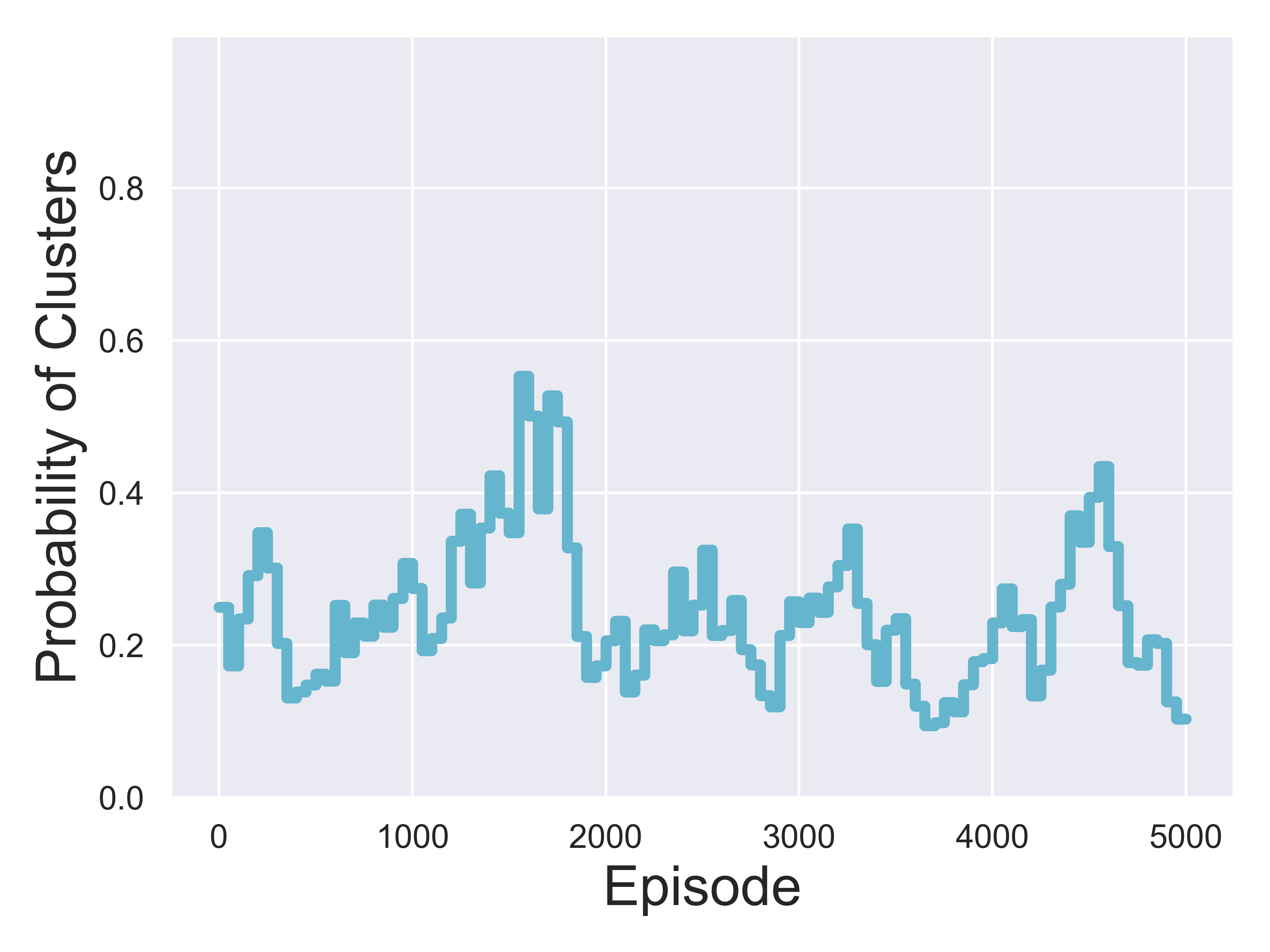}\label{fig_res_cls2}}
    \subfigure[Fourth cluster ($N_{\text{ss}}=3$)]{\includegraphics[width=.24\textwidth]{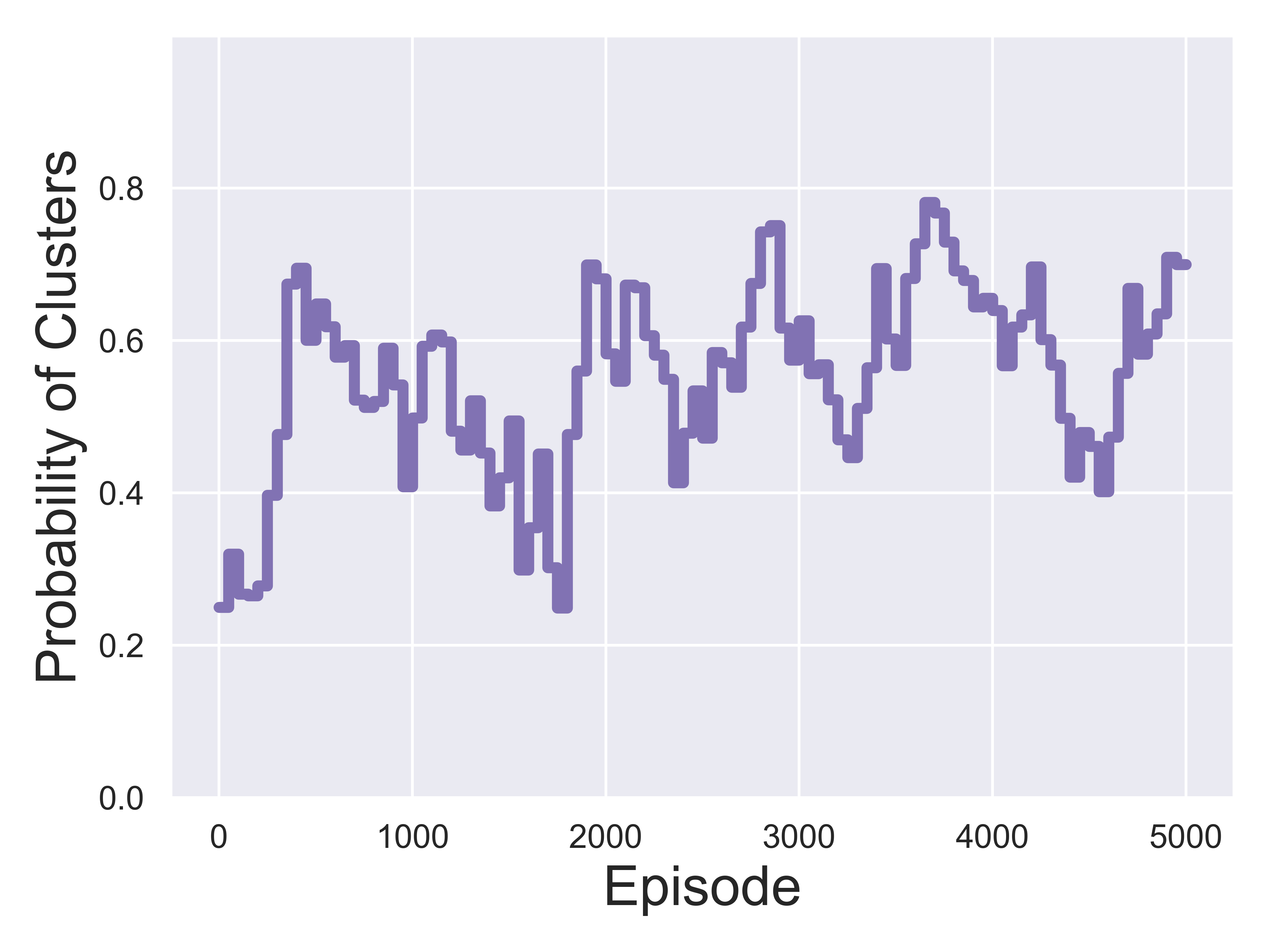}\label{fig_res_cls3}}
\caption{Weight updates of four clusters in BiMAB during the training process.
}
\label{fig:res_clusters}
\end{figure*}

\begin{figure*}[!htbp] 
\centering
    \subfigure[Arms in first cluster.]{\includegraphics[width=.24\textwidth]{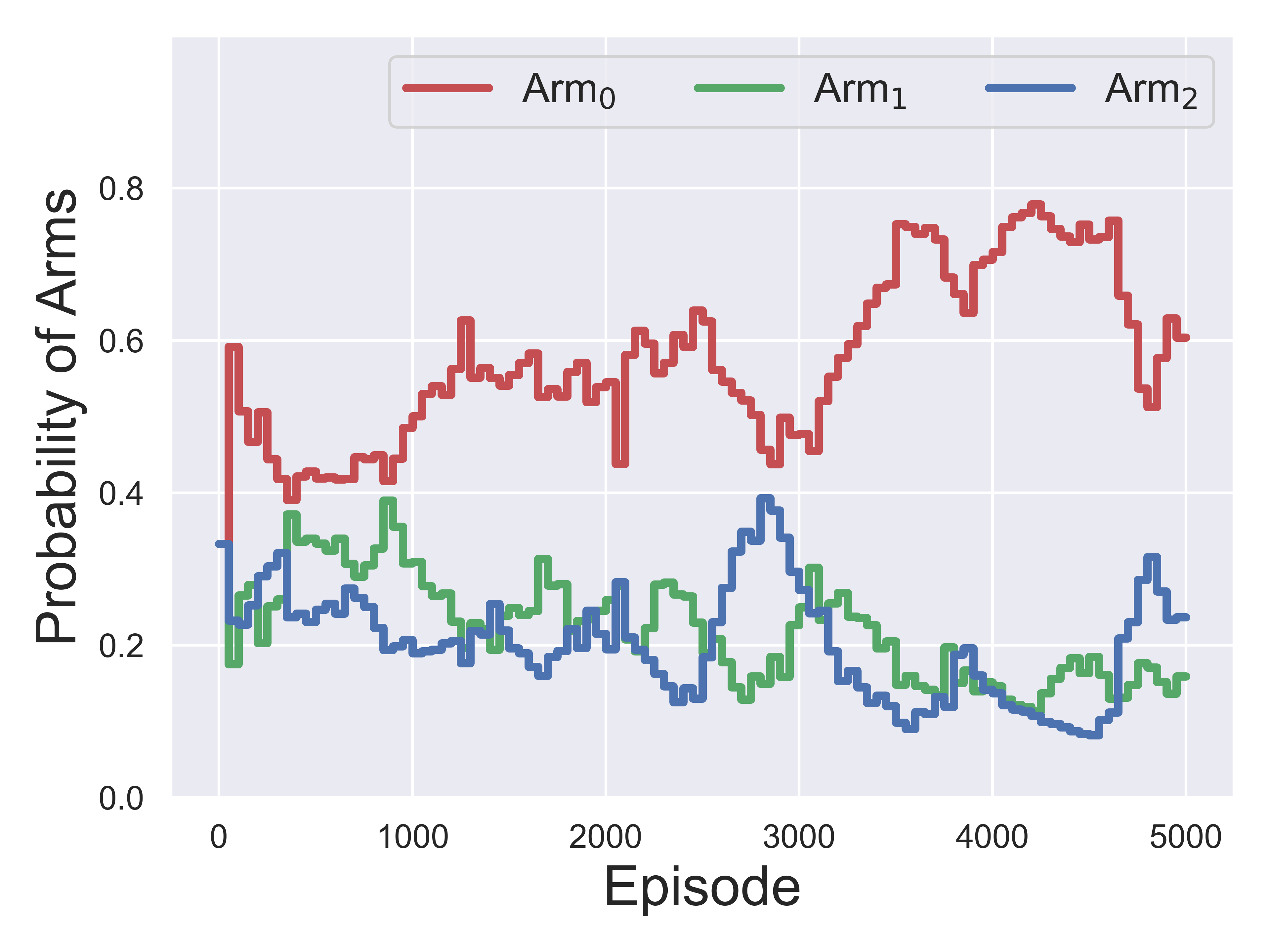}\label{fig_res_arm0}}
    \subfigure[Arms in second cluster.]{\includegraphics[width=.24\textwidth]{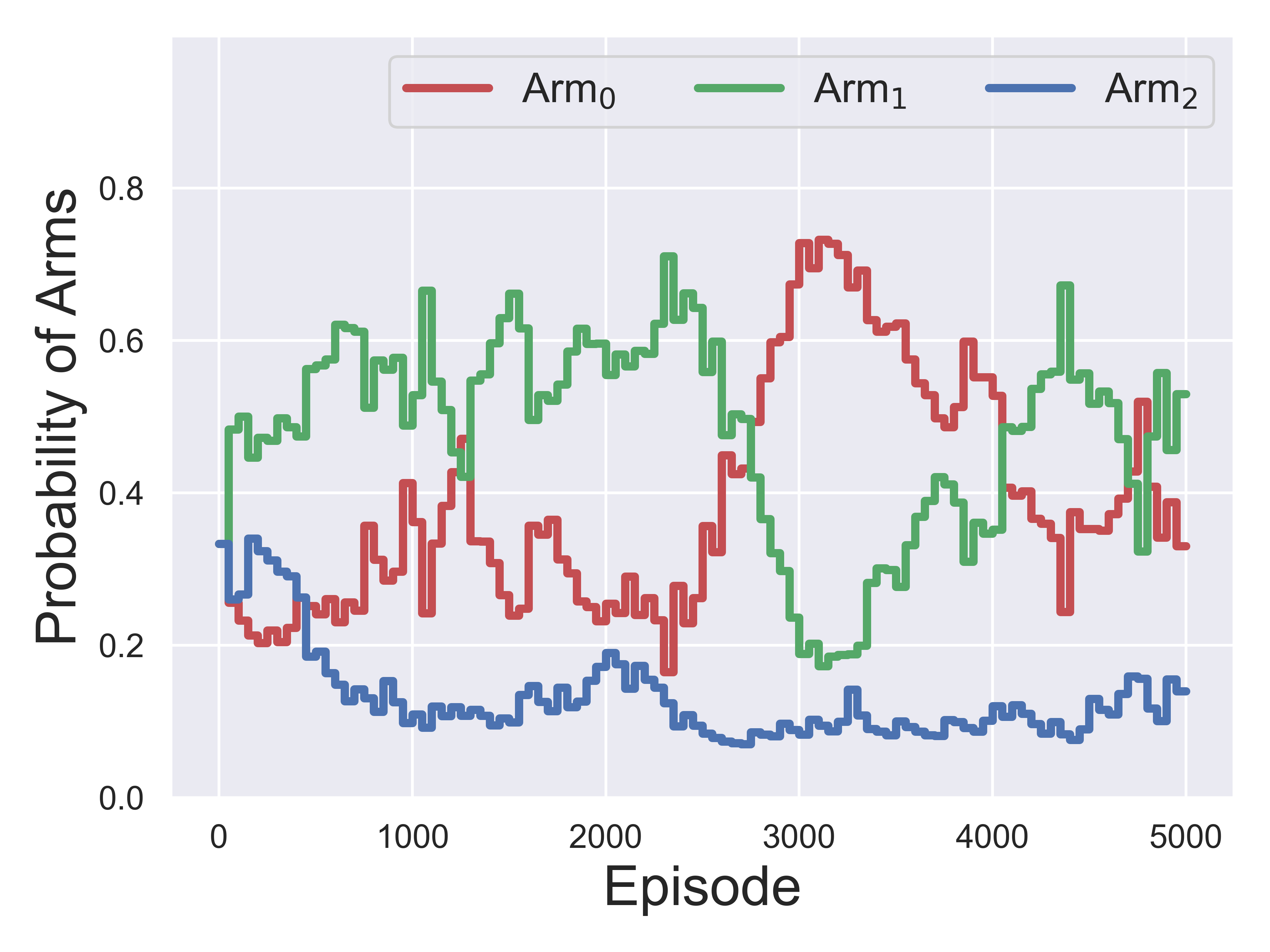}\label{fig_res_arm1}}
    \subfigure[Arms in third cluster.]{\includegraphics[width=.24\textwidth]{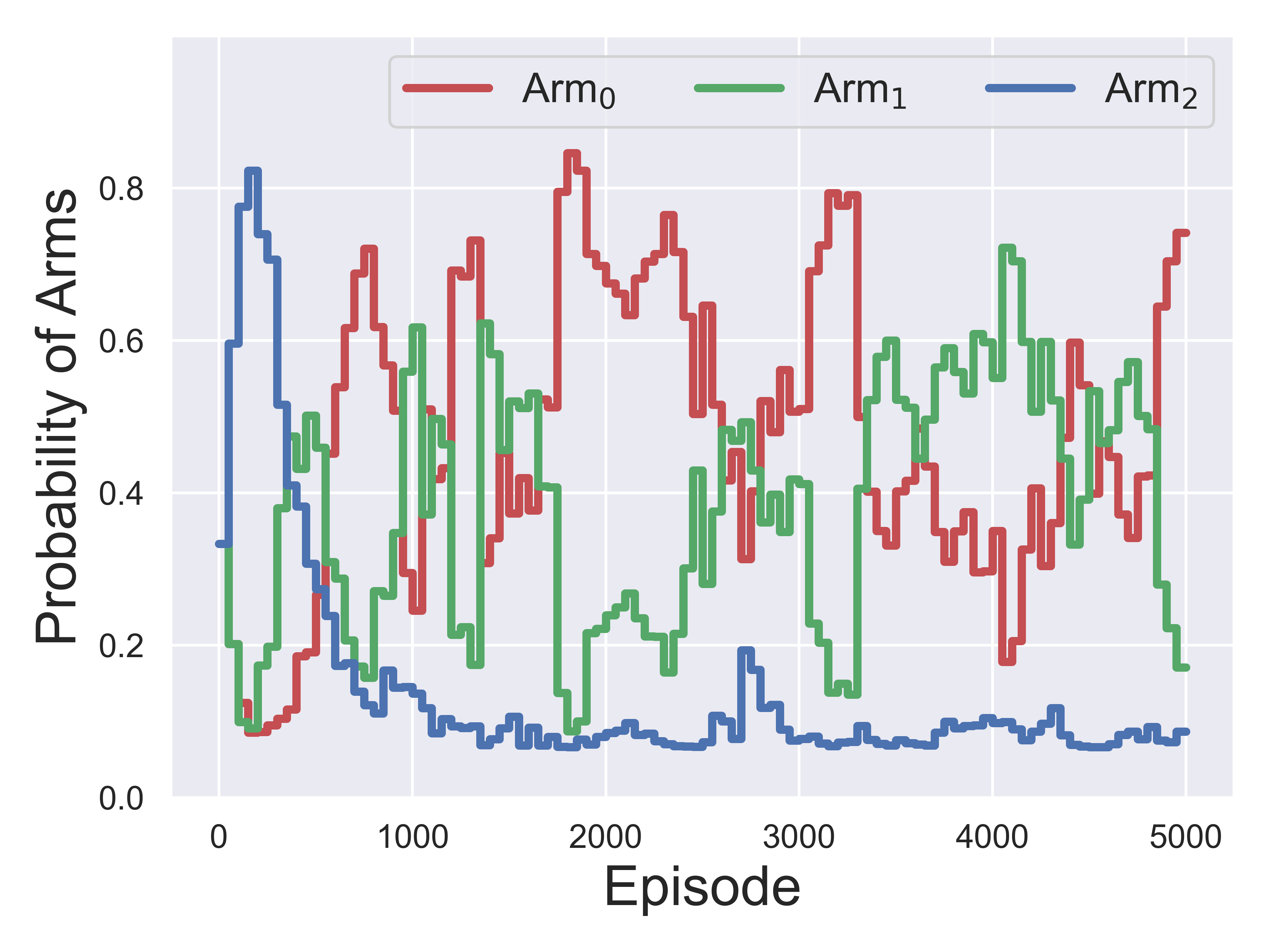}\label{fig_res_arm2}}
    \subfigure[Arms in fourth cluster.]{\includegraphics[width=.24\textwidth]{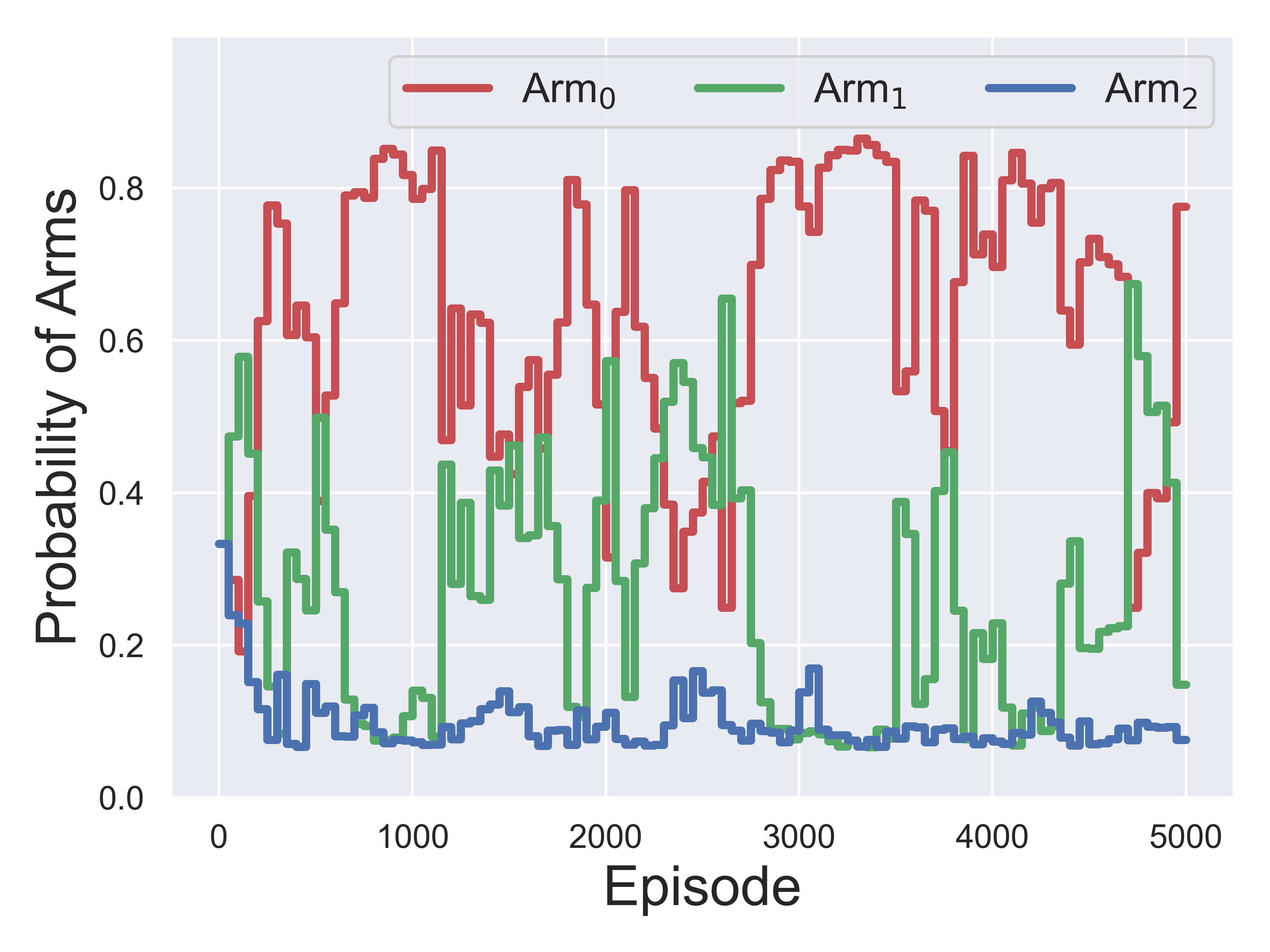}\label{fig_res_arm3}}
\caption{Weight updates of three arms in respective clusters during the training process.
}
\label{fig:res_arms}
\end{figure*}

\begin{table*}[!htbp] 
\renewcommand{\arraystretch}{1}
\scriptsize
\centering
\caption{Sampled times of clusters and arms within BiMAB
}
\label{table_2lane_BiMAB}
\begin{tabular}{c|ccc|ccc|ccc|ccc}
\hline
$N_{\text{ss}}$       & \multicolumn{3}{c|}{0}    & \multicolumn{3}{c|}{1}      & \multicolumn{3}{c|}{2}    & \multicolumn{3}{c}{3}    \\ \hline
Sampled Times & \multicolumn{3}{c|}{456}  & \multicolumn{3}{c|}{536} & \multicolumn{3}{c|}{1291} & \multicolumn{3}{c}{2718}  \\ \hline
$N_{\text{sc}}$       & 0        & 1      & 2     & 0        & 1      & 2       & 0        & 1      & 2     & 0        & 1      & 2   \\ \hline
Sampled Times & 268      & 90     & 98    & 193      & 278    & 65   & 627      & 489    & 175   & 1788     & 666    & 264 \\ \hline
\end{tabular}
\end{table*}

\subsection{Dual-Lane Unsignalized Intersections}
\label{exp:dual-lane}

We select a dual-lane intersection from the Town05 map for the deployment, where the traffic lights are deactivated to create an unsignalized intersection scenario. In this case, the maximum number of SVs is set to $N_{\text{sv}}^{\max} = 3$. 
The start and target positions of the EV and SVs are randomly generated while complying with traffic rules, as described in Section \ref{Section2}.

\subsubsection{Training Results}

The trend of weight changes for each cluster (subset) and arm (sub-curriculum) of the BiMAB within the proposed method during the training process is illustrated in Figs. \ref{fig:res_clusters}-\ref{fig:res_arms} and Table \ref{table_2lane_BiMAB}. According to Table \ref{table_2lane_BiMAB}, we can find that the sampling times of the four clusters increase with the number of SVs throughout the entire training process. This is because as the number of SVs increases, the difficulty of the task also increases, 
requiring more training episodes for RL agents to learn and update strategies. 
This process is automatically assigned through real-time evaluation of the RL policy during training by BiMAB. 
Within specific clusters, the sampling frequency of arms corresponding to left-turn and go-straight tasks surpasses that of right-turn tasks, and the difference in sampling frequency increases significantly with the increase in the number of SVs. This phenomenon occurs because the first two tasks are more challenging than right-turn tasks, thereby resulting in such allocation results.

More details about the weight evolutions in BiMAB are illustrated in Figs. \ref{fig:res_clusters}-\ref{fig:res_arms}. 
It is observable that as training progresses, the weights in the first layer of BiMAB gradually shift from cluster $N_{ss}=0$ to cluster $N_{ss}=3$. This indicates that BiMAB dynamically assesses the learning progress of the strategy and executes a timely shift in the curriculum. 
Specifically, as the performance of the RL policy improves, the BiMAB allocates increasingly challenging training episodes to the RL agent. This strategic allocation enables the RL agent to master complex driving skills step by step, thereby allowing it to achieve higher rewards in more challenging scenarios. 
Furthermore, because right-turn tasks are relatively simpler compared to the other two types of crossing tasks, the weights associated with the arm for right-turn tasks decrease as training progresses across all clusters. In contrast, the weights for left-turn and go-straight tasks exhibit diverse trends of change across different clusters, yet both types of tasks experience periods of high weight in various stages of training. This pattern is also evident from Table \ref{table_2lane_BiMAB}, which shows that right-turn tasks have the fewest training instances. Despite the potential collision points being equal for both left-turn and go-straight tasks, the left-turn tasks involve a longer distance through the central region of the unsignalized intersections, leading to a higher potential frequency of interaction with SVs. Consequently, BiMAB allocates a higher number of instances to left-turn tasks compared to go-straight tasks. 

\subsubsection{Performance Evaluation}

To quantitatively compare the performance of the BiM-ACPPO with the baseline methods, 
we conduct tests on RL policies trained by all methods in the dual-lane unsignalized intersection with the number of SVs ranging from 0 to 3. Each policy undergoes 100 repeated tests in each type of crossing task. 
The test results are summarized in Table \ref{table:test_res_2lane}.

\begin{table*}[!htbp]
\renewcommand{\arraystretch}{1}
\scriptsize
\centering
\caption{Performance comparison at dual-lane unsignalized intersections among different methods.
}
\label{table:test_res_2lane}
\resizebox{\linewidth}{!}{
\begin{tabular}{cl|ccc|ccc|ccc|ccc}
\hline
\multicolumn{2}{c|}{\multirow{2}{*}{Methods}}            & \multicolumn{3}{c|}{$N_{\text{sv}}$=0}             & \multicolumn{3}{c|}{$N_{\text{sv}}$=1}             & \multicolumn{3}{c|}{$N_{\text{sv}}$=2}             & \multicolumn{3}{c}{$N_{\text{sv}}$=3}              \\ \cline{3-14} 
\multicolumn{2}{c|}{}                                    & S(\%) & C(\%) & TO(\%) & S(\%) & C(\%) & TO(\%) & S(\%) & C(\%) & TO(\%) & S(\%) & C(\%) & TO(\%) \\ \hline
\multicolumn{1}{c|}{\multirow{3}{*}{Fixed PPO}} & Left Turn   & 98         & 0         & 2            & 76         & 24         & 0            & 64         & 36         & 0            & 36         & 64         & 0            \\
\multicolumn{1}{c|}{}                      & Go Straight & 92         & 0         & 8            & 79         & 18         & 3            & 72         & 24         & 4            & 48         & 52         & 0            \\
\multicolumn{1}{c|}{}                      & Right Turn  & 100         & 0         & 0            & 98         & 1         & 1            & 98         & 2         & 0            & 96         & 4         & 0            \\ \hline
\multicolumn{1}{c|}{\multirow{3}{*}{Manual CPPO}} & Left Turn   & 100         & 0         & 0            & 83         & 17         & 0            & 60         & 40         & 0            & 54         & 46         & 0            \\
\multicolumn{1}{c|}{}                      & Go Straight & 100         & 0         & 0         & 84         & 16            & 0         & 78         & 22            & 0         & 66         & 34            & 0            \\
\multicolumn{1}{c|}{}                      & Right Turn  & 100         & 0         & 0            & 100         & 0         & 0            & 98         & 2         & 0            & 97         & 3         & 0            \\ \hline
\multicolumn{1}{c|}{\multirow{3}{*}{Random CPPO}} & Left Turn   & 100         & 0         & 0            & 82         & 18         & 0            & 66         & 34         & 0            & 44         & 56         & 0            \\
\multicolumn{1}{c|}{}                      & Go Straight & 100         & 0         & 0            & 84         & 16         & 0            & 70         & 30         & 0            & 62         & 38         & 0            \\
\multicolumn{1}{c|}{}                      & Right Turn  & 100         & 0         & 0            & 100         & 0         & 0            & 96         & 1 4        & 0            & 84         & 16         & 0            \\ \hline
\multicolumn{1}{c|}{\multirow{3}{*}{RD-ACPPO}} & Left Turn   & 100         & 0         & 0            & 86         & 14         & 0            & 79         & 21         & 0            & 69         & 31         & 0            \\
\multicolumn{1}{c|}{}                      & Go Straight & 100         & 0         & 0            & 86         & 14         & 0            & 77         & 23         & 0            & 73         & 27         & 0            \\
\multicolumn{1}{c|}{}                      & Right Turn  & 100         & 0         & 0            & 100         & 0         & 0            & 100         & 0         & 0            & 98         & 2         & 0            \\ \hline
\multicolumn{1}{c|}{\multirow{3}{*}{\textbf{BiM-ACPPO}}} & Left Turn   & \textbf{100}         & \textbf{0}         & \textbf{0}           & \textbf{94}         & \textbf{6}         & \textbf{0}            & \textbf{85}         & \textbf{15}         & \textbf{0}            & \textbf{80}         & \textbf{20}         & \textbf{0}            \\
\multicolumn{1}{c|}{}                      & Go Straight & \textbf{100}         & \textbf{0}         & \textbf{0}           & \textbf{93}         & \textbf{7}         & \textbf{0}            & \textbf{87}         & \textbf{13}         & \textbf{0}             & \textbf{84}         & \textbf{16}         & \textbf{0}            \\
\multicolumn{1}{c|}{}                      & Right Turn  & \textbf{100}         & \textbf{0}         & \textbf{0}            & \textbf{100}         & \textbf{0}         & \textbf{0}           & \textbf{100}         & \textbf{0}         & \textbf{0}            & \textbf{100}         & \textbf{0}         & \textbf{0}            \\ \hline
\end{tabular}
}
\begin{tablenotes}[flushleft] 
\item \textit{Note:} S, C, and TO represent success rate, collision rate, and timeout rate, respectively.
\end{tablenotes}
\end{table*}

According to this table, it is evident that the proposed approach achieves the highest success rate in all testing tasks and settings, although the success rate decreases as the number of SVs increases. This result indicates the effectiveness of BiMAB in improving the training outcomes of the RL policy. 
Besides, we can find that the performance of RL policies trained across various scenario settings generally surpasses that of Fixed PPO, with RL policies utilizing curriculum learning outperforming Random PPO. Furthermore, the success rate of RL policies that automatically allocate curricula based solely on the number of SVs is higher than that of manually assigned curricula, while lower than that of proposed BiMAB.  
It is noteworthy that the success rate of the RL policy based on BiMAB significantly exceeds that of RD-ACPPO. This highlights the efficacy of leveraging the structured nature of unsignalized intersection tasks for curriculum modeling, as well as the effectiveness of the Algorithm \ref{ALG_BiMAB}. 

Meanwhile, the success rates of BiM-ACPPO in left-turn and go-straight tasks are similar across different numbers of SVs, while they are lower than that of right-turn tasks, which corroborates the analysis presented in Section \ref{Section3-BiMAB} regarding the three types of crossing tasks at unsignalized intersections.   
This discrepancy is due to the significantly higher number of potential collision points in the latter two tasks compared to right-turn tasks, necessitating more interactions with SVs that exhibit uncertain driving intentions. The complexity of these interactive processes exponentially increases with the number of SVs. 
Additionally, the Fixed PPO policy experiences time-out results in testing scenarios where the number of SVs ranged from 0 to 2. 
This reveals that directly training the RL agent in complex environments could lead to poor generalization to unseen scenarios, while curriculum learning techniques can assist in enhancing the generalization performance of the RL policy across different scenario settings.

\begin{figure*}[!htbp] 
\centering
    \subfigure[Left-turn task. The EV is initialized in the right lane, and its target position is set to the left lane of the left region.]{\includegraphics[width=0.85\textwidth]{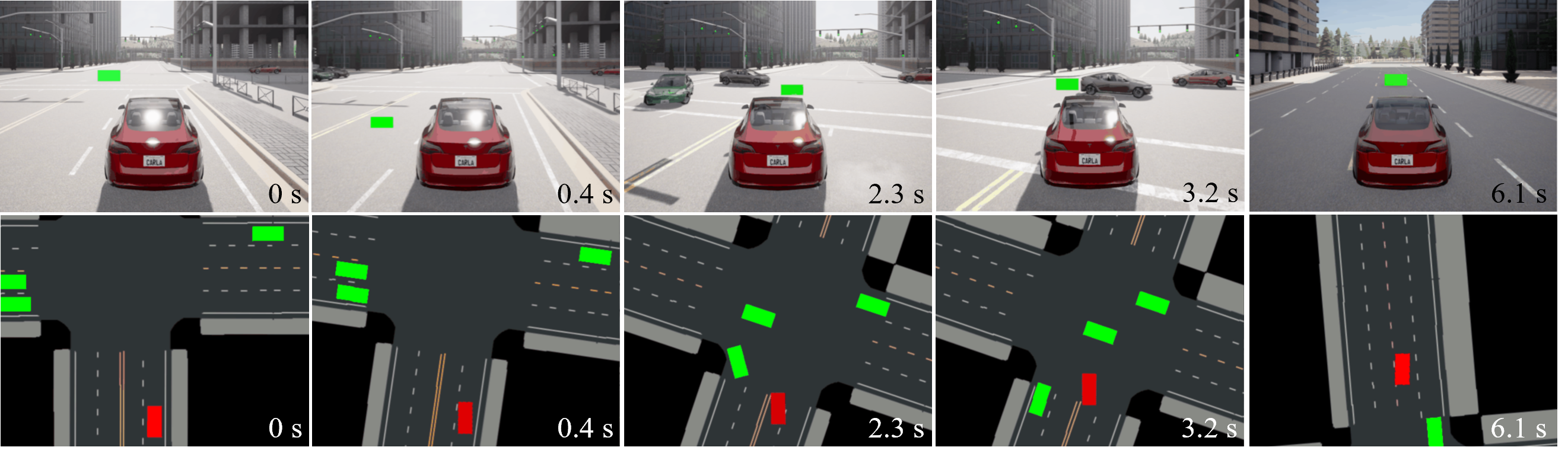}\label{fig_demo_left}}
    \subfigure[Go-straight task. The EV is initialized in the left lane, and its target position is set to the left lane of the upper region.]{\includegraphics[width=0.85\textwidth]{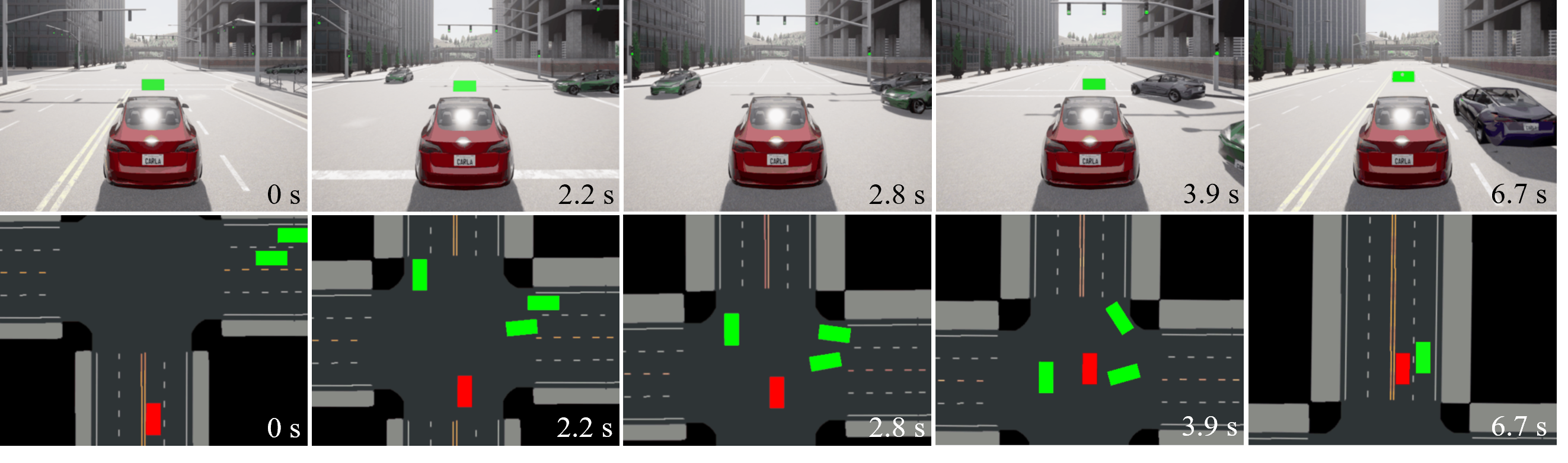}\label{fig_demo_straight}}
    \subfigure[Right-turn task. The EV is initialized in the left lane, and its target position is set to the left lane of the right region.]{\includegraphics[width=0.85\textwidth]{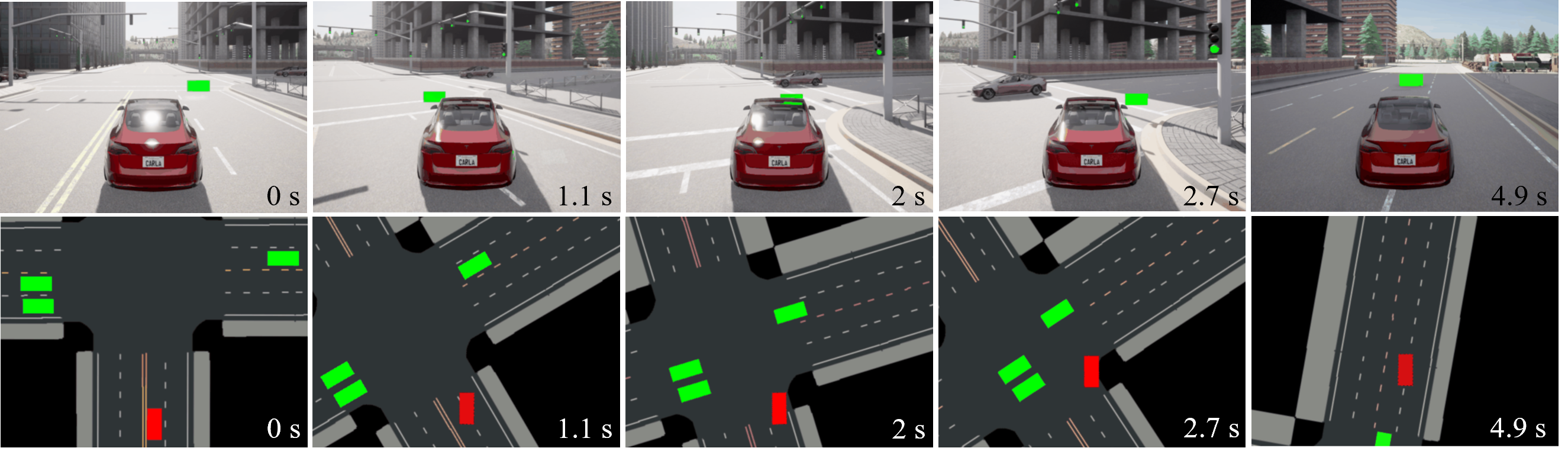}\label{fig_demo_right}}
\caption{Key frames of the three demonstrations with our method in an unsignalized intersection within CARLA. The upper and lower side of the sub-figures shows the third-person views of the EV and the bird-eye views. 
The green rectangles in the third-person sub-figures denote the intermediate point determined by the RL policy. The red rectangle and the green rectangles in the bird-eye sub-figures represent the EV and SVs, respectively. 
}
\label{fig:demo_3task}
\end{figure*}

Among all testing results attained by the BiM-ACPPO approach, we pick up one result from each of the three different crossing tasks at the unsignalized intersection for demonstration. The snapshots of these three examples are presented in Fig. \ref{fig:demo_3task}.

\begin{itemize}
    \item \textbf{Left-turn task}: 
    At $0.4$ s, the RL policy outputs an intermediate waypoint in the left lane to guide the EV to change lanes in preparation for the left-turn crossing task. Subsequently, at $2.3$ s, the RL policy detects an SV approaching from the left region to execute a go-straight crossing task without showing any intention to yield. Therefore, the RL policy selects the nearest waypoint and a low reference speed to perform the yield behavior. Once the SV has passed and is no longer in front of the EV, the RL policy then chooses a distant waypoint along with a high reference speed to rapidly direct the EV towards the target region at $3.2$ s. Ultimately, the EV finishes the left-turn task under the guidance of the RL policy. 
    \item \textbf{Go-straight task}: 
    Because there are no SVs in the central region, the RL policy chooses a distant waypoint and a high reference speed as an intermediate point to guide the EV for rapid transit from $0$ s to $2.2$ s. At $2.8$ s, the RL policy observes two SVs entering the central region from the right region, posing a potential collision risk. Thus, it selects the nearest waypoint and a low reference speed to prevent potential collisions. Between $2.8$ s and $3.9$ s, the two SVs from the right region, with the SV in the right lane indicating an intent to turn right and the SV in the left lane showing an intention to yield, prompt the RL policy to choose a distant waypoint to swiftly guide the EV through the central region. Finally, the EV completes the go-straight task. 
    \item \textbf{Right-turn task}: 
    The EV is initialized in the left lane, with its destination being the left lane of the right region of the unsignalized intersection. Between $0$ s and $1.1$ s, the RL policy selects a waypoint in the right lane to guide the EV to change lanes in preparation for a right turn. From $2$ s to $2.7$ s, the RL policy observes the SVs that come from the left region are at a sufficiently safe distance, which poses no potential collision risk. Therefore, it outputs a sub-nearest waypoint to direct the EV through a tight turn with a small radius into the right lane of the right region of the unsignalized intersection. At $4.9$ s, the EV successfully enters the target region under the guidance of the RL policy and changes lanes to reach its target location. 
\end{itemize}

\vspace{-0.3cm}

\subsection{Generalization Ability Into New Scenarios}

To demonstrate the generalization ability of the proposed framework, we deploy the RL policies obtained in \ref{exp:dual-lane} to conduct tests in two new driving scenarios for validation.

\subsubsection{Zero-Shot Generalization Ability Into Single-Lane Unsignalized Intersections}

In this experiment, we choose a single-lane unsignalized intersection in the map Town03 of the CARLA simulator as the testing scenario. Other test settings are similar to dual-lane unsignalized intersections in Section \ref{exp:dual-lane}. 
During testing, RL policies trained by Fixed PPO, Manual CPPO, and Random CPPO are almost incapable of completing intersection crossing tasks in this new scenario. Therefore, we present the results of 100 repeated tests on RD-ACPPO and the proposed framework, which are shown in Table \ref{table:test_robust_1lane}. 

\begin{table*}[t] 
\renewcommand{\arraystretch}{1}
\scriptsize
\centering
\caption{Performance comparison at single-lane unsignalized intersections among different methods.
}
\label{table:test_robust_1lane}
\resizebox{\linewidth}{!}{
\begin{tabular}{cl|ccc|ccc|ccc|ccc}
\hline
\multicolumn{2}{c|}{\multirow{2}{*}{Methods}}            & \multicolumn{3}{c|}{$N_{\text{sv}}$=0}             & \multicolumn{3}{c|}{$N_{\text{sv}}$=1}             & \multicolumn{3}{c|}{$N_{\text{sv}}$=2}             & \multicolumn{3}{c}{$N_{\text{sv}}$=3}              \\ \cline{3-14} 
\multicolumn{2}{c|}{}                                    & S(\%) & C(\%) & TO(\%) & S(\%) & C(\%) & TO(\%) & S(\%) & C(\%) & TO(\%) & S(\%) & C(\%) & TO(\%) \\ \hline
\multicolumn{1}{c|}{\multirow{3}{*}{RD-ACPPO}} & Left Turn   & 100         & 0         & 0            & 77         & 23         & 0            & 75         & 25         & 0            & 56         & 44         & 0            \\
\multicolumn{1}{c|}{}                      & Go Straight & 100         & 0         & 0            & 75         & 25         & 0            & 65         & 35         & 0            & 53         & 47         & 0            \\
\multicolumn{1}{c|}{}                      & Right Turn  & 100         & 0         & 0            & 88         & 12         & 0            &  84        & 16         & 0            & 82         & 18         & 0            \\ \hline
\multicolumn{1}{c|}{\multirow{3}{*}{\textbf{BiM-ACPPO}}} & Left Turn   & \textbf{100}         & \textbf{0}         & \textbf{0}           & \textbf{91}         & \textbf{9}         & \textbf{0}            & \textbf{85}         & \textbf{15}         & \textbf{0}            & \textbf{80}         & \textbf{20}         & \textbf{0}            \\
\multicolumn{1}{c|}{}                      & Go Straight & \textbf{100}         & \textbf{0}         & \textbf{0}           & \textbf{96}         & \textbf{4}         & \textbf{0}            & \textbf{89}         & \textbf{11}         & \textbf{0}             & \textbf{82}         & \textbf{18}         & \textbf{0}            \\
\multicolumn{1}{c|}{}                      & Right Turn  & \textbf{100}         & \textbf{0}         & \textbf{0}            & \textbf{100}         & \textbf{0}         & \textbf{0}           & \textbf{100}         & \textbf{0}         & \textbf{0}            & \textbf{97}         & \textbf{3}         & \textbf{0}            \\ \hline
\end{tabular}
}
\begin{tablenotes}[flushleft] 
\item \textit{Note:} S, C, and TO represent success rate, collision rate, and timeout rate, respectively.
\end{tablenotes}
\end{table*}

Since the BiMAB automatically modulates the curriculum in line with the training progress of the RL policy, it can effectively guide the RL agent towards an enhanced exploration of the environment. This enables the trained RL policy to gain an improved comprehension of the environment, handle uncertainty in the environment well, and demonstrate enhanced generalization capabilities. 
Additionally, comprehensive utilization of the inherent task structure at unsignalized intersections within the BiMAB framework facilitates a superior success rate in the novel single-lane unsignalized intersection scenarios compared to RD-ACPPO. 

\begin{figure*}[!t] 
    \centering   
    \includegraphics[trim=0 0cm 0 0cm, width=0.85\linewidth]{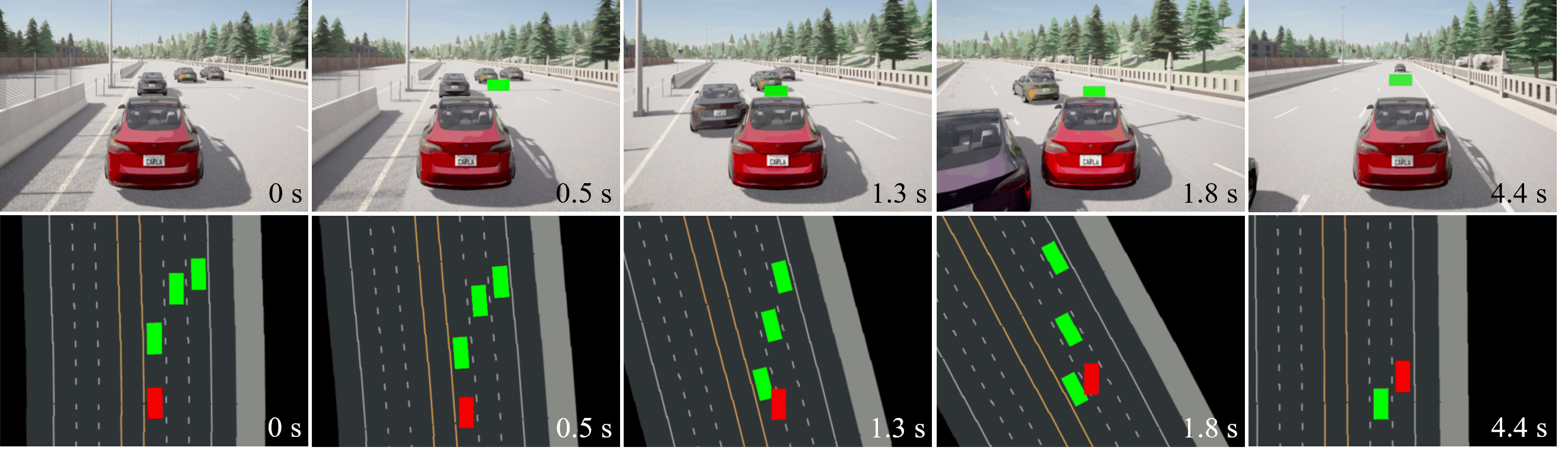}
   \caption{Key frames of the demonstration with our method in the overtaking task. 
   SVs in the left lane and middle lane are set to low-speed mode. 
   }
    \label{fig:demo_overtake}
\end{figure*}

\subsubsection{Few-Shot Generalization Ability Into Overtaking Driving Scenarios}
To further validate the generalization of the proposed method, we also conduct tests on the overtaking task in an urban driving scenario. 
Here, we select the long outer ring road with 3 lanes in the map Town05 of the CARLA simulator as the testbed. 
The RL policy trained by the BiM-ACPPO approach is fine-tuned and then applied in the constructed overtaking scenario. 
Here, we consider 3 SVs with the Tesla Model 3. All SVs are in built-in autopilot mode and initialized in random positions. 
During the fine-tuning stage, a random number of SVs ($N_{sv}=0,1,2$) is set to a low-speed mode with different speed limits. During the testing phase, several SVs are randomly set to low-speed mode to block traffic to verify the overtaking ability of the RL policy. 
Several SVs are randomly set to low-speed mode to block traffic to verify the overtaking ability of the RL policy. 
We conduct 100 repeated tests on the proposed framework, which achieves a success rate of 97$\%$. The results indicate the commendable generalization ability of the proposed method in overtaking tasks.

Among all the results attained by the proposed method, we pick up one result for demonstration, which is shown in Fig. \ref{fig:demo_overtake}. 
In this example, the EV is initialized in the left lane, with the SVs in the left and middle lanes set to low-speed mode. Specifically, the speed limits for the SVs on the left and in the middle are set to $10\%$ and $20\%$ of the normal speed limit, respectively. 
Between $0.5$ s and $1.3$ s, the RL policy observes that the SV ahead is moving slower than the EV, prompting it to change lanes to the right at a low reference speed to guide the EV into the middle lane to achieve safe and efficient driving behavior. At $1.3$ s, the RL agent notices that the SVs in the left and middle lanes showed no intention of accelerating, while an SV in the right lane is accelerating. Therefore, between $1.3$ s and $1.8$ s, the RL policy guides the EV to change lanes to the right one. Finally, at $4.4$ s, the EV completes the overtaking task and continues driving forward.

Based on the experimental results, we can observe that the proposed framework demonstrates commendable generalization ability across different test scenarios, which can be attributed to the design of the overall framework. Firstly, the automatic curriculum learning mechanism based on BiMAB enables the RL policy to interact with the environment from easy to difficult during training, thereby enhancing the training efficiency and the performance of the trained policy. Secondly, the introduction of the hierarchical RL structure along with a multi-discrete action space allows the trained RL policy to achieve a high success rate and perform well in scenarios not encountered during training.

\vspace{-0.1cm}

\section{Conclusion}

In this work, we present a novel BiMAB-based hierarchical RL framework for interaction-aware self-driving at unsignalized intersections, which aims to address the uncertainties arising from multi-modal behaviors and varying number of SVs. 
The training problem of the RL policy in target scenarios is modeled as a bilevel curriculum learning task and is addressed by the proposed BiMAB algorithm. 
Experimental results demonstrate that our BiMAB could automatically evaluate the learning progress and adaptively assign suitable curricula for the RL agent throughout the training process. 
Our approach achieves the highest success rates compared to all baseline methods in statistical testing experiments. 
Furthermore, the proposed method showcases commendable generalization ability in new scenarios.
Demonstrations in different driving scenarios highlight the interaction-aware ability of the proposed method with multi-modal SVs. 
\vspace{-0.1cm}

\bibliographystyle{IEEEtran}
\bibliography{pzqbib}

\vfill

\end{document}